\documentclass{article} 
\usepackage{iclr2027_conference,times}

\usepackage{amsmath,amsfonts,bm}

\def\1{\bm{1}}

\DeclareMathAlphabet{\mathsfit}{\encodingdefault}{\sfdefault}{m}{sl}
\SetMathAlphabet{\mathsfit}{bold}{\encodingdefault}{\sfdefault}{bx}{n}

\usepackage{hyperref}
\usepackage{url}
\usepackage{booktabs}       
\usepackage{multicol, multirow}
\usepackage{amsfonts}       
\usepackage{amsmath,amssymb}
\usepackage{nicefrac}       
\usepackage{microtype}      
\usepackage{xcolor}         
\usepackage{caption}

\usepackage{siunitx}

\usepackage{tikz}
\usetikzlibrary{arrows.meta, positioning, calc, backgrounds, decorations.pathreplacing}
\usepackage{pgfplots}
\usepgfplotslibrary{groupplots}
\usetikzlibrary{shapes.geometric}
\usetikzlibrary{shapes.misc}     

\definecolor{ink}{HTML}{000000}
\definecolor{s1col}{HTML}{2563A8}
\definecolor{s2col}{HTML}{C0533A}
\definecolor{encol}{HTML}{2E7D52}
\definecolor{encol}{HTML}{2E7D52}
\definecolor{incol}{HTML}{5B4B8A}
\definecolor{outcol}{HTML}{A0692A}
\definecolor{aggcol}{HTML}{3A8FA8}

\definecolor{bg1}{HTML}{F7F5F0} 
\definecolor{bg2}{HTML}{F2F4F7} 
\definecolor{bg3}{HTML}{EEF2F5} 

\usepackage{algorithm,algpseudocode}

\tikzset{
  box/.style 2 args={
    rectangle, rounded corners=5pt,
    minimum width=#1, minimum height=1.1cm,
    draw=#2, line width=1.2pt, fill=#2!15,
    font=\sffamily\small\bfseries, text=ink, align=center
  },
  arr/.style={-{Stealth[length=5pt,width=4pt]}, line width=1.4pt, #1},
  line/.style={line width=1.4pt, #1},
  darr/.style={-{Stealth[length=3.5pt,width=3pt]}, line width=0.85pt, dashed, #1},
  lbl/.style={font=\sffamily\fontsize{6.8}{8}\selectfont, align=center},
  sub/.style={font=\sffamily\fontsize{6}{7.2}\selectfont\itshape, align=center},
  gn/.style={circle, draw=ink, fill=white, line width=0.8pt, minimum size=10pt, inner sep=0pt, font=\sffamily\fontsize{5}{5}\selectfont},
  co/.style={rectangle, rounded corners=3pt, draw=#1, fill=#1!12, inner sep=3.5pt, font=\sffamily\fontsize{6.2}{7.5}\selectfont, align=center, text=ink, line width=0.8pt},
}

\title{Two-Stage Learned Decomposition for \\ Scalable Routing on Multigraphs}

\author{
\textbf{Filip Rydin\thanks{Correspondence to: \texttt{filipry@chalmers.se}}$\:\:^1$, Morteza Haghir Chehreghani$^{1, 2}$ \& Balázs Kulcsár$^1$} \\[0.5em]
$^1$Chalmers University of Technology, $^2$University of Gothenburg
}

\iclrfinalcopy 
\begin{document}

\maketitle

\begin{abstract}
Most neural methods for Vehicle Routing Problems (VRPs) are limited to Euclidean settings or simple graphs. In this work, we instead consider \emph{multigraphs}, where parallel edges represent distinct travel options with varying trade-offs (e.g., distance vs.~time). Multigraphs are highly relevant in practice, yet few neural methods are designed for them, and those that do exist face major scalability issues. We address these scalability issues with Node-Edge Policy Factorization (NEPF), which splits the routing policy into a node permutation stage and an edge selection stage. To enable the decomposition, we introduce a pre-encoding edge aggregation scheme and a non-autoregressive architecture for the edge stage, as well as a hierarchical reinforcement learning method to train the stages jointly. Our experiments across six VRP variants demonstrate that NEPF trains and runs up to orders of magnitude faster than prior neural multigraph methods and scales to considerably larger instances, while matching or improving on their solution quality\footnote{Our code will be made publicly available upon paper acceptance.}.
\end{abstract}

\section{Introduction}

Learning-based methods for Vehicle Routing Problems (VRPs) are emerging as efficient alternatives to classical methods, such as exact solvers and meta-heuristics \citep{ZhouLearningforRouting}. Their overall aim is to obtain high-quality solutions in real time with minimal engineering effort to adapt to new problem variants. Recently, several works have even begun exploring cross-problem generalization, paving the way for future VRP foundation models \citep{drakulic2025goal, URS}. 

However, flexibility in input representation remains a key limitation. Most neural methods operate either on Euclidean instances \citep{Kool2018AttentionLT, kwon_pomo_2020} or on graphs with a single precomputed edge between each node pair \citep{kwon_matrix_2021, drakulic2023bqnco}. Real-world transportation networks often present multiple alternative paths between locations with differing attributes (e.g., distance, travel time and reliability), naturally leading to a multigraph formulation. Multigraph routing, i.e., jointly optimizing path selection and node visit order, often reduces costs substantially compared to fixing paths in advance and then routing in simple graphs \citep{ben_ticha_empirical_2017}. Despite this, neural approaches for VRPs on multigraphs remain in their early stages.

Only one recent work, \citet{rydin2026beyond}, has explored this setting, but their methods face two problems. First, they maintain the full $\mathcal{O}(M N^2)$ edge-level multigraph representation (where $N$ is the number of nodes and $M$ is parallel edges per node pair) throughout encoding, greatly increasing memory footprint compared to node-based $\mathcal{O}(N)$ representations. Second, their decoding operates directly on this dense edge-level representation and is tightly coupled to the multigraph structure. Taken together, these issues prevent scaling to instances where the node count $N$ and the number of parallel edges $M$ grow simultaneously, which is a regime of high practical interest \citep{ben_ticha_empirical_2017}. These problems also hinder wider adoption of the multigraph framework, for instance in foundation models, where compatibility with standard node-based architectures is desirable.

We show that a classical decomposition strategy from operations research \citep{garaix_2010} can be effectively integrated into modern neural routing models and that it provides a powerful principle for \emph{scalable} learning on multigraphs. Concretely, we factorize the routing solution into a node permutation and a sequence of edge choices, and model these components with separate policies trained jointly with hierarchical reinforcement learning. We ensure efficiency in the node permutation stage by proposing a pre-encoding aggregation mechanism that summarizes each set of parallel edges into a compact representation. To tackle the edge selection stage, we introduce a light but effective non-autoregressive architecture that operates on the sequence of edge sets.
Our contributions are:
\vspace{-0.15cm}
\begin{itemize}
    \item Conceptually, we propose Node-Edge Policy Factorization (NEPF), a new formulation for learning-based routing on multigraphs that decouples high-level node decisions from edge-level choices, confining the dense $\mathcal{O}(MN^2)$ representation to shallow pre- and post-processing components.
    \item Methodologically, to enable the two-stage decomposition, we introduce a pre-encoding edge aggregation scheme that compresses parallel edges into a latent representation, significantly reducing memory and computational cost. We also design a lightweight non-autoregressive architecture for the edge selection stage and train both stages jointly using a hierarchical reinforcement learning framework.
    \item Experimentally, we demonstrate that NEPF is up to orders of magnitude faster to train and run than existing neural multigraph methods and scales to substantially larger instances, while matching or improving on solution quality across six single-objective and multi-objective problems. We also show that the framework is compatible with a range of node-based backbones, making it a promising building block for more general routing models.
\end{itemize}

\section{Related Work}

\vspace{-0.15cm}
\paragraph{Learning for routing on richer graph representations.}
While early work \citep{vinyals2015pointer, bello2017neural, Kool2018AttentionLT, kwon_pomo_2020} focused almost exclusively on Euclidean VRP instances, more recent approaches consider richer and more realistic representations. These include asymmetric graphs \citep{kwon_matrix_2021, drakulic2023bqnco, yi2026radar}, graphs with dynamic costs \citep{yang2025neural} and real-world instances with multi-modal travel costs \citep{son2026towards}.

To the best of our knowledge, the only prior work on multigraph VRPs is \citet{rydin2026beyond}, which introduced two methods: GMS-EB and GMS-DH. Both methods require the full $\mathcal{O}(MN^2)$ multigraph representation throughout encoding, causing major scalability issues. Moreover, GMS-EB features edge-based autoregressive decoding, incurring $\mathcal{O}(MN^4)$ complexity, while GMS-DH requires pre-pruning the entire graph before constructing routes, degrading performance when parallel edges are numerous or edge selection is difficult. 

In contrast, we propose a two-stage factorization that first commits to a node permutation and only then selects edges. This avoids maintaining the dense multigraph during most processing, yielding significantly faster training and inference while retaining strong performance across diverse problem settings. Our node-centric approach also ensures compatibility with many neural backbones, enabling easier integration with, for example, architectures for constraint handling \citep{BiLearningToHandle, chen_looking_2024, bi2026towards} or cross-problem generalization \citep{drakulic2025goal, URS}.

\vspace{-0.15cm}
\paragraph{Classical methods for multigraph VRPs.} In the operations research literature, multigraphs are widely used to capture the presence of multiple distinct paths between locations in road networks \citep{Lai2016, ben_ticha_empirical_2017, benticha_2019, tikani_2021}. \citet{garaix_2010} introduced the two-step decomposition we build on, namely to first fix the node permutation and then solve the edge selection problem, which is known as the \emph{Fixed Sequence Arc Selection Problem} (FSASP). While they solve the FSASP with dynamic programming, we replace both stages with end-to-end learning, enabling GPU acceleration, cross-problem adaptation, and handling of cases (e.g., negative resource consumption) that complicate classical methods. See Appendix~\ref{app: FSASP_motivation} for extended discussion.

Appendix~\ref{app: related_work} reviews additional related work, focusing on the learning for routing literature and existing decomposition-based methods therein. We also note that our approach is widely applicable to combinatorial optimization beyond routing, which we discuss further in Appendix~\ref{app:beyond-routing}.

\section{Problem Formulation and Notation}
\label{sec: notation}
The goal in a multigraph routing problem is to find the feasible route $r\in \mathcal{R}$ in a multigraph $G$ that minimizes a cost $C(r)$. We assume that $G$ is directed, with nodes $V$ and edges $E$. It has edge attributes $e_l,\, l\in E$ (e.g., distance or travel time) and, depending on the VRP at hand, node attributes $n_u,\,u\in V$ (e.g., node demand). Denote the connecting edge set between nodes $(u, v) \in V^2$ as $E_{uv}$. 

A route $r$ is fully described by a sequence of edges, but we consider an alternative formulation. In our case, a solution is described by a node sequence $\boldsymbol{\pi} = (\pi_1, \dots, \pi_T)$, together with a sequence $\boldsymbol{\epsilon} = (\epsilon_1, \dots, \epsilon_{T-1})$ of \emph{connecting} edges, such that $\epsilon_t \in E_{\pi_t\pi_{t+1}}$. Formally, the fixed sequence arc selection problem is to determine the optimal $\boldsymbol{\epsilon}$ for a fixed $\boldsymbol{\pi}$. This problem is NP-hard in many cases \citep{garaix_2010}, being equivalent to a multidimensional multiple choice knapsack problem.

\vspace{-0.15cm}
\paragraph{Multi-objective optimization.} Multigraphs are highly relevant in the Multi-Objective (MO) setting, as each objective typically corresponds to a different edge attribute (consider for instance the $m$-objective TSP with $m$ edge distances). As such, following \citet{rydin2026beyond}, we heavily emphasize MO problems. The goal is to find the \emph{Pareto set} of routes. Redefine $C: \mathcal{R} \to \mathbb{R}^m$ as a multi-objective route cost, where $C_i(r)$ denotes the cost of $r$ under objective $i$. Formally, the Pareto set is the set 
\begin{equation}
    \text{PS} := \left\{ r \in \mathcal{R} \;\middle|\; \nexists r' \in \mathcal{R} : 
    \left(C_i(r') \leq C_i(r) \;\forall i \right) 
    \wedge 
    \left( \exists i :\; C_i(r') < C_i(r) \right) 
    \right\}.
\end{equation}
That is, a solution is in the Pareto set if no alternative solution is at least as good in all objectives and strictly better in one. The \emph{Pareto front} is the corresponding set in the objective space. 

To solve MO routing problems, we utilize \emph{Chebyshev scalarization}, which transforms an MO problem into several single-objective subproblems. Define a preference vector $\lambda \in \mathbb{R}^m$ such that $\lambda_i\geq 0\:\forall i$ and $\sum_i \lambda_i = 1$. Given the ideal value $z_i^*$ for objective $i$, the Chebyshev cost is defined as 
\begin{equation}
\label{eq: Chb}
    C_\lambda(r) := \max_i \{ \lambda_i |C_i(r) - z^*_i|\}.
\end{equation}
For a fixed $\lambda$, the Chebyshev subproblem is to solve $\min_r C_\lambda(r)$. Given a solution $r^*$ in the Pareto set, there always exists a corresponding preference for which $r^*$ solves the Chebyshev subproblem \citep{ProperEfficiency}. That is, by sweeping over the preferences, this scalarization can in principle recover the entire Pareto front. In contrast, naive linear scalarization, i.e., using the $\lambda$-weighted sum of objectives, cannot find solutions on non-convex parts of the front \citep{EhrgottMultiCriteria}. 

\section{Node-Edge Policy Factorization (NEPF) for Multigraph VRPs}
\label{sec: method}
Building on the classical decomposition by \citet{garaix_2010}, our framework is designed to first output $\boldsymbol{\pi}$ and then $\boldsymbol{\epsilon}$, yielding the complete route $r=(\boldsymbol{\pi}, \boldsymbol{\epsilon})$. Denote the joint policy $p_\theta(r \mid G)$. By the chain rule, we factorize this joint policy as 
\begin{equation}
    p_\theta(r \mid G) = p^{\text{node}}_{\theta_1}(\boldsymbol{\pi} \mid G)  \,p^{\text{edge}}_{\theta_2}(\boldsymbol{\epsilon} \mid \boldsymbol{\pi}, G).
\end{equation}
Our first stage, referred to below as the node permutation stage, samples $\boldsymbol{\pi} \sim p^{\smash{\text{node}}}_{\theta_1}$, while the second stage solves the FSASP to obtain $\boldsymbol{\epsilon} \sim p^{\smash{\text{edge}}}_{\theta_2}$. We couple the stages through joint hierarchical training. The full setup is visualized in Figure~\ref{fig: method}.

\begin{figure}
    \centering
    \resizebox{\textwidth}{!}{
        \input{figures/method.tikz}
    }
    \caption{Overview of the proposed node-edge policy factorization for multigraph VRPs. The model first aggregates parallel edges into a compact representation, then generates a node permutation and finally selects edges conditioned on the node sequence. This avoids propagating the full $\mathcal{O}(M N^2 d)$ embedded multigraph through most of the network, while retaining strong performance. The backbone can be exchanged for any architecture that processes distance matrices and outputs node permutations, which makes the framework compatible with emerging foundation models, such as GOAL \citep{drakulic2025goal} and URS \citep{URS}.
    }
    \label{fig: method}
    \vspace{-0.25cm}
\end{figure}

\subsection{Node Permutation Stage}

\vspace{-0.15cm}
\paragraph{Pre-encoding aggregation.} In our framework, rather than processing the entire dense multigraph in the encoding stage like \citet{rydin2026beyond}, we first compute a latent $d$-dimensional distance matrix $D \in \mathbb{R}^{N \times N \times d}$ using a small pre-encoding module. This compression reduces the memory footprint of the full $\mathcal{O}(M N^2)$ multigraph representation. After linear embedding of edge attributes: $g_l = W_g\, e_l$, we obtain a latent distance vector per node-pair with Deep Sets \citep{DeepSets}: 
\begin{equation}
\label{eq: pre-encoding}
    D_{uv} = \rho \Big ( \sum_{l \in E_{uv}} \phi(g_l) \Big ), \quad (u, v) \in V^2,
\end{equation}
where $\rho$ and $\phi$ are Multilayer Perceptrons (MLPs). We note that this aggregation is permutation-invariant with respect to the edge ordering. The latent distance matrix captures the metric relationship for each node-pair, without explicitly retaining each edge as a unique entity.

\vspace{-0.15cm}
\paragraph{Encoding.} The encoder maps the latent distance matrix $D$ to node encodings $h_u \in \mathbb{R}^d, \,u \in V$. As architecture, we utilize the Graph Edge Attention Network (GREAT) of \citet{lischkagreat} together with a small number of transformer layers \citep{VaswaniTsf} to obtain expressive node embeddings. This is the same base encoder design as GMS-DH. Unlike many alternative architectures, it is capable of producing expressive enough embeddings for the multi-objective asymmetric graph scenario \citep{rydin2026beyond}.

More specifically, the first $L_1$ layers of the encoder are Node-Based (NB) GREAT layers. A single such layer is organized as a transformer layer and includes an attention sublayer, a feedforward layer, skip connections and normalization (batchnorm in the original implementation, which we replace with layernorm), all applied to the edge embeddings. The attention sublayer propagates edge embeddings $D_{uv} \mapsto D'_{uv}$ by forming temporary node features $x_u$, according to
\begin{align}
\label{eq: GREAT}
    & x_u = \sum_{v \in \mathcal{N}_u} \Big (\alpha'_{uv} W' D_{uv} \| \alpha''_{uv} W'' D_{vu} \Big ),\\
\label{eq: GREAT2}
    & D'_{uv} = W''' \big (x_u \| x_v \big ),
\end{align}
where $\|$ denotes concatenation, $\mathcal{N}_u$ is the neighborhood of node $u$ and $\alpha', \alpha''$ are learnable attention scores computed via a gating-mechanism. In the last GREAT layer, we only apply \eqref{eq: GREAT}, refraining from projecting back the node features to edges with \eqref{eq: GREAT2}. These are then passed through $L_2$ transformer layers, yielding node encodings $h_u, \,u \in V$. Note that GREAT layers are completely edge-based. Thus, if the original graph has node attributes, we concatenate them with edge attributes before \eqref{eq: pre-encoding}.

\vspace{-0.15cm}
\paragraph{Decoding.} The decoder autoregressively calculates probabilities $p^\text{node}_{\theta_1}(\pi_t \nobreak\mid \nobreak \pi_{1:t-1}, G)$ for the next node, given context from the partial route $\pi_{1:t-1}$ and encoded nodes $h_u, \,u \in V$. To parameterize this policy, we utilize a Multi-Pointer (MP) decoder \citep{jin_pointerformer_2023}. We generally refer to the original paper for details. However, we make three modifications to fit this decoder into our setting. 

First, we adapt the $-\text{cost}(u, v)$ term in the score calculation (which penalizes expensive node transitions $u \to v$, improving convergence) to the multigraph setting. We set $\text{cost}(u, v)$ slightly differently depending on the setting (see Appendix~\ref{app: hyper}), but in general we use the (scalarized) cost of the cheapest edge between $u$ and $v$. 

Second, we weigh this term with a learnable scalar $\beta$. This yields superior performance compared to the default choice $\beta = 1$, see Appendix~\ref{app: ablations} for an ablation study.

Third, we solve the problem of partial observability that is introduced by selecting edge traversal after node permutation. To clarify, with our decomposition we deprive the model of full state information for certain problems. For instance, consider time-dependent VRPs, where current time is determined by traversed edges. Often, it is beneficial performance-wise to include such state information in the decoder query. Consequently, to remedy this problem, we propose an auxiliary recurrent neural network-based state estimator that restores accurate state estimates $\hat{s}_t$ given the partial route $\pi_{1:t-1}$. See Appendix~\ref{app: state} for full details.

\vspace{-0.15cm}
\paragraph{Flexible backbone.} While we utilize a GREAT-based encoder and autoregressive multi-pointer decoder for the node permutation stage, we emphasize that our framework is backbone-agnostic. In fact, we only require it to be a function $f_\text{NP} : \mathbb{R}^{N \times N \times d} \times \mathbb{R}^{N \times d} \to \Pi$, mapping the learned distance matrix $D$ as well as optional embedded node features $z_u \in \mathbb{R}^d, \,u \in V$ to a node permutation $\boldsymbol{\pi} \in \Pi$. Consequently, our framework is highly compatible with existing routing models and in Section~\ref{sec: cross-architecture}, we show promising results with various backbones.

\vspace{-0.15cm}
\subsection{Fixed Sequence Arc Selection Problem Stage}
In the second stage, the input is a sequence $\mathcal{E}_{\boldsymbol{\pi}} := (E_1, \dots, E_{T-1})$ of edge sets $E_t := E_{\pi_t\pi_{t+1}}$ and the task is to select one edge $\epsilon_t \in E_t$ per set. To start, for each edge $l$, we concatenate $e_l$ with (potential) node attributes $n_u$ of the end-point node $u$, obtaining augmented features $e'_l$. These are mapped to $\mathbb{R}^{\smash{d'}}$ with a linear layer: $f_l = W_f\, e'_l$. We then form set-level embeddings via mean pooling:
\begin{equation}
    f^\text{pooled}_t = \frac{1}{|E_t|} \sum_{l \in E_t} f_l, \quad t =1, \dots, T-1.
\end{equation}
To ensure edge selections are informed by full route context, we mix the sequence of aggregated embeddings with a bi-directional recurrent neural network (thereby avoiding attention, which would scale quadratically with $T$). Concretely, mixed embeddings are formed with a BiLSTM and a skip connection according to 
\begin{equation}
\label{eq: mixing}
    f^{\text{mixed}}_t = \text{BiLSTM}(f^\text{pooled}_1, \dots, f^\text{pooled}_{T-1})_t + f^\text{pooled}_t.
\end{equation}
Subsequently, to calculate edge scores, we set keys to the embeddings $k_l = f_l$ and queries for each position to the mixed embeddings $q_t = f^{\text{mixed}}_t$. We form the score $\alpha_l$ for edge $l \in E_t$ with a multi-pointer mechanism:
\begin{equation}
\label{eq: score}
    \alpha_l = \frac{1}{H} \sum_{h=1}^{H} \frac{1}{\sqrt{d'}} (W^q_h q_t)^T (W_h^k k_l) - \widetilde{\beta}\,\text{cost}(l), \quad l \in E_t,
\end{equation}
where $H$ is the number of heads, $\widetilde{\beta}$ is a learnable parameter and $\text{cost}$ is a problem-specific cost function that penalizes expensive edges. To obtain per-step normalized edge probabilities, we apply tanh-clipping and softmax within each edge set $E_t$, $t=1,\dots, T-1$. Edges in different positions are then sampled independently in parallel. This non-autoregressive one-shot sampling allows us to efficiently obtain multiple candidates $\boldsymbol{\epsilon}\sim p_{\theta_2}^{\smash{\text{edge}}}(\cdot \nobreak\mid\nobreak \boldsymbol{\pi}, G)$ for evaluation. 

In summary, the edge stage applies the composition: $\text{pooling} \rightarrow \text{BiLSTM mixing} \rightarrow \text{scoring}$, which maps the edge-set sequence $\mathcal{E}_{\boldsymbol{\pi}}$ to a per-position distribution. While this FSASP setup is rather shallow, it is robust enough for our purpose. It consistently performs well across problems and the lightweight architecture results in little latency overhead compared to heuristically selecting edges independently per position. 

\vspace{-0.15cm}
\subsection{Hierarchical Training Setup}
\label{sec: hierarchical_training}
We propose a hierarchical training algorithm to train the node permutation stage and the FSASP stage jointly. The overall goal is to maximize
\begin{equation}
    J(\theta) = \mathbb{E}_{r \sim p_\theta,\,G \sim \mathcal{G}} \big [ R(r) \big ] = \mathbb{E}_{\boldsymbol{\epsilon} \sim p^\text{edge}_{\theta_2},\, \boldsymbol{\pi} \sim p^\text{node}_{\theta_1},\,G \sim \mathcal{G}} \big [R(\boldsymbol{\pi}, \boldsymbol{\epsilon}) \big ],
\end{equation}
where $R(\boldsymbol{\pi}, \boldsymbol{\epsilon}) = - C(\boldsymbol{\pi}, \boldsymbol{\epsilon})$ is the negative cost of the route $r = (\boldsymbol{\pi}, \boldsymbol{\epsilon})$. We estimate the gradient of this expectation and use it in gradient ascent. With our probability factorization and the REINFORCE estimator \citep{Reinforce}, we obtain
\begin{equation}
\label{eq: grad}
    \nabla_\theta J(\theta) = \mathbb{E}_{\boldsymbol{\epsilon} \sim p^\text{edge}_{\theta_2},\, \boldsymbol{\pi} \sim p^\text{node}_{\theta_1},\,G \sim \mathcal{G}} \Big [ R(\boldsymbol{\pi}, \boldsymbol{\epsilon})\,\big ( \nabla_{\theta_1}\log p^\text{node}_{\theta_1}(\boldsymbol{\pi} \mid G) + \nabla_{\theta_2}\log p^\text{edge}_{\theta_2}(\boldsymbol{\epsilon}\mid \boldsymbol{\pi}, G) \big ) \Big ].
\end{equation}
To compute a Monte Carlo approximation, we generate $B$ multigraph instances per batch and $K_1$ POMO node permutations \citep{kwon_pomo_2020} with differing starting nodes. Additionally, for each node permutation and instance, we sample $K_2$ edge selections. As already argued, this has negligible runtime effect as long as $K_2$ remains moderate. Setting $R^*(\boldsymbol{\pi}_{ij}) = \max_{k=1, \dots, K_2} R(\boldsymbol{\pi}_{ij}, \boldsymbol{\epsilon}_{ijk})$ as the reward for a node permutation, we define the baselines
\begin{equation}
    b^\text{node}(G_i) = \frac{1}{K_1} \sum_{j=1}^{K_1} R^*(\boldsymbol{\pi}_{ij}), \quad b^\text{edge}(\boldsymbol{\pi}_{ij}, G_i) = \frac{1}{K_2} \sum_{k=1}^{K_2} R(\boldsymbol{\pi}_{ij}, \boldsymbol{\epsilon}_{ijk}).
\end{equation}
We then estimate the gradient \eqref{eq: grad} according to 
\begin{align}
\begin{split}
\label{eq: reinforce}
    \nabla_\theta J(\theta) \approx 
    &\frac{1}{B K_1} \sum_{i=1}^B \sum_{j=1}^{K_1} \big ( R^*(\boldsymbol{\pi}_{ij}) - b^\text{node}(G_i) \big ) \nabla_{\theta_1} \log p^\text{node}_{\theta_1}(\boldsymbol{\pi}_{ij} \mid G_i) + \\
    &\frac{1}{B K_1 K_2} \sum_{i=1}^B \sum_{j=1}^{K_1} \sum_{k=1}^{K_2} \big ( R(\boldsymbol{\pi}_{ij}, \boldsymbol{\epsilon}_{ijk}) - b^\text{edge}(\boldsymbol{\pi}_{ij}, G_i) \big ) \nabla_{\theta_2} \log p^\text{edge}_{\theta_2}(\boldsymbol{\epsilon}_{ijk} \mid \boldsymbol{\pi}_{ij}, G_i).
    \raisetag{4.3\baselineskip}
\end{split}
\end{align}
The complete per-batch procedure is summarized in Algorithm~\ref{alg: training} (Appendix~\ref{app: pseudo-code}). We refer to this as hierarchical RL because the edge policy $p^\text{edge}$ optimizes local decisions conditioned on a fixed node sequence, while the node policy $p^\text{node}$ is trained on the best achievable downstream reward $R^*$ over multiple edge samples. This induces a two-level optimization scheme, where the lower-level edge policy informs the training signal of the higher-level node policy. 

\subsection{Multi-Objective Preference Conditioning}

In the multi-objective setting, following \citet{lin_pareto_2022}, we use MLP hypernetworks to condition decoding on a preference vector. This allows a single model to recover the full Pareto front without retraining and avoids recomputing node encodings per preference. We use separate MLPs for the two stages, mapping the preference $\lambda$ to the multi-pointer scoring weights. We also adopt the Chebyshev reward $R_\lambda(r) = -C_\lambda(r)$ for training and sample one preference per batch when estimating \eqref{eq: reinforce}.

\section{Experimental Results}
\label{sec: results}

We evaluate NEPF across six VRP variants, demonstrating robust performance, significant efficiency gains and improved scalability over state-of-the-art multigraph methods.

\vspace{-0.15cm}
\paragraph{Problems.} In the single-objective setting we benchmark on a Resource Constrained TSP (RCTSP) and a multi-resource Orienteering Problem (OP). In the multi-objective setting we use the Multi-Objective TSP (MOTSP), CVRP (MOCVRP) and TSP with Time-Windows (MOTSPTW) of \citet{rydin2026beyond}, together with a bi-objective OP (MOOP). All problems have multiple edge features, with different edges offering varying trade-offs, hence the multigraph framework is appropriate. Instances are generated from the FLEX$x$ and FIX$x$ distributions of \citet{rydin2026beyond}, where FLEX$x$ has a variable number of edges per node pair (at most $x$) with uncorrelated features and FIX$x$ has exactly $x$ edges with negatively correlated features. We benchmark with $x=2,5$ and $100$ nodes, and complement this with more realistic instances (Appendix~\ref{app: rw}) and real-world instances (Appendix~\ref{app: real-world-instances}). Further details on problems and instance generation are given in Appendix~\ref{app: problems}.

\vspace{-0.15cm}
\paragraph{Baselines.} As classical single-objective baseline, we use Gurobi on mixed-integer linear program formulations. For the multi-objective problems we combine state-of-the-art single-objective heuristics with preference-based scalarization and multigraph pre-pruning, namely LKH \citep{LKH3} for the MOTSP, HGS \citep{HGS} for the MOCVRP and Google OR-tools \citep{ortools2019} for both. We additionally include simpler problem-specific heuristics, detailed in Appendix~\ref{app: problems}. As learning-based baselines we use optimized re-implementations of GMS-EB and GMS-DH \citep{rydin2026beyond}, which are, to our knowledge, the only existing neural methods for multigraph VRPs and hence the current neural state-of-the-art. GMS was originally designed for multi-objective problems, but we adapt both variants to the single-objective setting by removing the preference-conditioning.

\vspace{-0.15cm}
\paragraph{Metrics.} For the single-objective problems, we report the objective value averaged over $10\,000$ random instances. Similarly, we report the Hypervolume (HV) \citep{HV} for the multi-objective problems, averaged over $200$ instances. In both cases, we also report the gap compared to the best method and the total inference time. 

\vspace{-0.15cm}
\paragraph{Model settings.} All neural models are trained for 200 epochs of $100\,000$ random instances on a single NVIDIA Tesla A40 GPU (\SI{48}{GB}), with NEPF hyperparameters in Appendix~\ref{app: hyper} and GMS hyperparameters taken from \citet{rydin2026beyond}. Classical methods run on an Intel Xeon Gold 6338 CPU with parallelization across 32 processes for a fairer comparison. At test time we additionally report $\times 8$ scaling-based instance augmentation \citep{lischkagreat} and use half-precision inference for all neural models, including the baselines. In the multi-objective setting, all methods sweep across 101 preferences to form the Pareto front.

\begin{table*}[t]
\centering
\caption{Multi-objective problems. Note that HV is ``higher-is-better'' and normalized to the range $[0, 1]$. The best neural method is in \textbf{bold} and the second best \underline{underlined}, while $\star$ indicates statistically significant difference compared to the second best neural method via a Wilcoxon signed-rank test with $1\%$ significance.}
\vspace{-0.15cm}
\label{tab: MO}
{
\setlength{\tabcolsep}{0.08cm}
\renewcommand{\arraystretch}{0.8}
\fontsize{8}{11}\selectfont
\begin{tabular}{clcccccccccccc}
\toprule
&\multicolumn{1}{l|}{}                        & \multicolumn{3}{c|}{FLEX2-100}                                              & \multicolumn{3}{c|}{FLEX5-100}                          & \multicolumn{3}{c|}{FIX2-100} & \multicolumn{3}{c}{FIX5-100} \\
&\multicolumn{1}{l|}{}                  & HV & Gap & \multicolumn{1}{c|}{Time}   & HV & Gap & \multicolumn{1}{c|}{Time}   & HV & Gap & \multicolumn{1}{c|}{Time}   & HV & Gap & Time   \\ 
\midrule

\multirow{11}{*}{\rotatebox{90}{\textbf{MOTSP}}} 

&\multicolumn{1}{l|}{LKH}          
& 0.94 & 0.00\% & \multicolumn{1}{c|}{(7.0m)}  
& 0.97 & 0.00\% & \multicolumn{1}{c|}{(5.9m)}
& 0.95 & 0.00\% & \multicolumn{1}{c|}{(5.9m)} 
& 0.91 & 0.00\% & \multicolumn{1}{c}{(5.6m)} 
\\

&\multicolumn{1}{l|}{OR-tools}          
& 0.91 & 3.46\% & \multicolumn{1}{c|}{(4.8m)}  
& 0.96 & 1.60\% & \multicolumn{1}{c|}{(4.3m)}
& 0.93 & 2.25\% & \multicolumn{1}{c|}{(4.3m)} 
& 0.90 & 1.78\% & \multicolumn{1}{c}{(4.4m)} 
\\

&\multicolumn{1}{l|}{Nearest Neighbor}          
& 0.88 & 6.71\% & \multicolumn{1}{c|}{(25s)}  
& 0.94 & 3.33\% & \multicolumn{1}{c|}{(25s)}
& 0.91 & 4.23\% & \multicolumn{1}{c|}{(25s)} 
& 0.88 & 3.11\% & \multicolumn{1}{c}{(25s)} 
\\

\cmidrule(lr){2-14}

&\multicolumn{1}{l|}{GMS-EB}          
& 0.93 & 1.52\% & \multicolumn{1}{c|}{(3.2m)}  
& 0.96 & 0.94\% & \multicolumn{1}{c|}{(7.7m)}
& 0.94 & 1.12\% & \multicolumn{1}{c|}{(3.2m)} 
& 0.90 & 0.97\% & \multicolumn{1}{c}{(7.7m)} 
\\

&\multicolumn{1}{l|}{GMS-EB (aug)}          
& \underline{0.93} & \underline{1.27\%} & \multicolumn{1}{c|}{(25m)}  
& \underline{0.97} & \underline{0.77\%} & \multicolumn{1}{c|}{(1.0h)}
& \underline{0.94} & \underline{0.93\%} & \multicolumn{1}{c|}{(25m)} 
& \underline{0.90} & \underline{0.91\%} & \multicolumn{1}{c}{(1.0h)} 
\\

&\multicolumn{1}{l|}{GMS-DH}          
& 0.92 & 2.22\% & \multicolumn{1}{c|}{(34s)}  
& 0.95 & 1.98\% & \multicolumn{1}{c|}{(33s)}
& 0.93 & 2.44\% & \multicolumn{1}{c|}{(31s)} 
& 0.89 & 2.43\% & \multicolumn{1}{c}{(48s)} 
\\

&\multicolumn{1}{l|}{GMS-DH (aug)}          
& 0.93 & 1.84\% & \multicolumn{1}{c|}{(4.0m)}  
& 0.96 & 1.50\% & \multicolumn{1}{c|}{(4.2m)}
& 0.93 & 2.21\% & \multicolumn{1}{c|}{(4.0m)} 
& 0.89 & 2.33\% & \multicolumn{1}{c}{(6.2m)} 
\\

\cmidrule(lr){2-14}

&\multicolumn{1}{l|}{NEPF}          
& 0.93 & 1.63\% & \multicolumn{1}{c|}{(11s)}  
& 0.96 & 0.95\% & \multicolumn{1}{c|}{(11s)}
& 0.94 & 1.21\% & \multicolumn{1}{c|}{(11s)} 
& 0.90 & 0.96\% & \multicolumn{1}{c}{(11s)} 
\\

&\multicolumn{1}{l|}{NEPF (aug)}          
& \textbf{0.93$^\star$} & \textbf{1.14\%} & \multicolumn{1}{c|}{(1.4m)}  
& \textbf{0.97$^\star$} & \textbf{0.67\%} & \multicolumn{1}{c|}{(1.4m)}
& \textbf{0.94$^\star$} & \textbf{0.92\%} & \multicolumn{1}{c|}{(1.4m)} 
& \textbf{0.91$^\star$} & \textbf{0.66\%} & \multicolumn{1}{c}{(1.4m)} 
\\

\midrule\midrule

\multirow{11}{*}{\rotatebox{90}{\textbf{MOCVRP}}} 

&\multicolumn{1}{l|}{HGS}          
& 0.89 & 0.00\% & \multicolumn{1}{c|}{(22h)}  
& 0.95 & 0.00\% & \multicolumn{1}{c|}{(22h)}
& 0.91 & 0.00\% & \multicolumn{1}{c|}{(22h)} 
& 0.87 & 0.00\% & \multicolumn{1}{c}{(22h)} 
\\

&\multicolumn{1}{l|}{OR-tools}          
& 0.84 & 5.96\% & \multicolumn{1}{c|}{(32m)}  
& 0.92 & 2.79\% & \multicolumn{1}{c|}{(31m)}
& 0.88 & 3.80\% & \multicolumn{1}{c|}{(33m)} 
& 0.84 & 3.12\% & \multicolumn{1}{c}{(31m)} 
\\

&\multicolumn{1}{l|}{Nearest Neighbor}          
& 0.73 & 17.49\% & \multicolumn{1}{c|}{(1.0m)}  
& 0.86 & 9.18\% & \multicolumn{1}{c|}{(1.0m)}
& 0.81 & 10.70\% & \multicolumn{1}{c|}{(1.0m)} 
& 0.81 & 7.29\% & \multicolumn{1}{c}{(1.0m)} 
\\

\cmidrule(lr){2-14}

&\multicolumn{1}{l|}{GMS-EB}          
& 0.86 & 3.26\% & \multicolumn{1}{c|}{(4.3m)}  
& 0.93 & 1.67\% & \multicolumn{1}{c|}{(10m)}
& 0.89 & 2.19\% & \multicolumn{1}{c|}{(4.4m)} 
& 0.84 & 3.63\% & \multicolumn{1}{c}{(12m)} 
\\

&\multicolumn{1}{l|}{GMS-EB (aug)}          
& \underline{0.87} & \underline{2.63\%} & \multicolumn{1}{c|}{(34m)}  
& \underline{0.93} & \underline{1.38\%} & \multicolumn{1}{c|}{(1.8h)}
& \underline{0.89} & \underline{1.87\%} & \multicolumn{1}{c|}{(35m)} 
& \underline{0.84} & \underline{3.51\%} & \multicolumn{1}{c}{(2.1h)} 
\\

&\multicolumn{1}{l|}{GMS-DH}          
& 0.84 & 5.97\% & \multicolumn{1}{c|}{(36s)}  
& 0.89 & 6.24\% & \multicolumn{1}{c|}{(41s)}
& 0.87 & 5.10\% & \multicolumn{1}{c|}{(39s)} 
& 0.84 & 3.88\% & \multicolumn{1}{c}{(56s)} 
\\

&\multicolumn{1}{l|}{GMS-DH (aug)}          
& 0.84 & 5.08\% & \multicolumn{1}{c|}{(4.6m)}  
& 0.91 & 4.37\% & \multicolumn{1}{c|}{(5.2m)}
& 0.87 & 4.57\% & \multicolumn{1}{c|}{(4.9m)} 
& 0.84 & 3.98\% & \multicolumn{1}{c}{(7.9m)} 
\\

\cmidrule(lr){2-14}

&\multicolumn{1}{l|}{NEPF}          
& 0.86 & 2.90\% & \multicolumn{1}{c|}{(16s)}  
& 0.93 & 1.38\% & \multicolumn{1}{c|}{(16s)}
& 0.89 & 2.10\% & \multicolumn{1}{c|}{(15s)} 
& 0.86 & 1.65\% & \multicolumn{1}{c}{(16s)} 
\\

&\multicolumn{1}{l|}{NEPF (aug)}          
& \textbf{0.87$^\star$} & \textbf{2.24\%} & \multicolumn{1}{c|}{(1.8m)}  
& \textbf{0.94$^\star$} & \textbf{1.06\%} & \multicolumn{1}{c|}{(1.9m)}
& \textbf{0.90$^\star$} & \textbf{1.72\%} & \multicolumn{1}{c|}{(1.8m)} 
& \textbf{0.86$^\star$} & \textbf{1.40\%} & \multicolumn{1}{c}{(1.8m)} 
\\

\midrule \midrule

\multirow{9}{*}{\rotatebox{90}{\textbf{MOTSPTW}}} 

&\multicolumn{1}{l|}{Insertion}          
& 0.92 & 1.67\% & \multicolumn{1}{c|}{(20m)}  
& 0.96 & 0.00\% & \multicolumn{1}{c|}{(1.8h)}
& 0.92 & 2.92\% & \multicolumn{1}{c|}{(25m)} 
& 0.86 & 4.09\% & \multicolumn{1}{c}{(1.8h)} 
\\

\cmidrule(lr){2-14}

&\multicolumn{1}{l|}{GMS-EB}          
& 0.63 & 32.41\% & \multicolumn{1}{c|}{(3.2m)}  
& 0.57 & 40.72\% & \multicolumn{1}{c|}{(6.6m)}
& 0.64 & 32.40\% & \multicolumn{1}{c|}{(3.2m)} 
& 0.88 & 2.31\% & \multicolumn{1}{c}{(7.7m)} 
\\

&\multicolumn{1}{l|}{GMS-EB (aug)}          
& 0.67 & 28.20\% & \multicolumn{1}{c|}{(26m)}  
& 0.60 & 38.19\% & \multicolumn{1}{c|}{(1.0h)}
& 0.69 & 27.42\% & \multicolumn{1}{c|}{(26m)} 
& \underline{0.90} & \underline{0.36\%} & \multicolumn{1}{c}{(1.0h)} 
\\

&\multicolumn{1}{l|}{GMS-DH}          
& 0.69 & 25.66\% & \multicolumn{1}{c|}{(32s)}  
& 0.56 & 41.76\% & \multicolumn{1}{c|}{(36s)}
& 0.95 & 0.63\% & \multicolumn{1}{c|}{(34s)} 
& 0.86 & 4.71\% & \multicolumn{1}{c}{(51s)} 
\\

&\multicolumn{1}{l|}{GMS-DH (aug)}          
& \underline{0.76} & \underline{18.52\%} & \multicolumn{1}{c|}{(4.6m)}  
& \underline{0.63} & \underline{35.10\%} & \multicolumn{1}{c|}{(5.2m)}
& \textbf{0.95$^\star$} & \textbf{0.00\%} & \multicolumn{1}{c|}{(4.9m)} 
& 0.89 & 0.83\% & \multicolumn{1}{c}{(7.2m)} 
\\

\cmidrule(lr){2-14}

&\multicolumn{1}{l|}{NEPF}          
& 0.93 & 0.70\% & \multicolumn{1}{c|}{(17s)}  
& 0.94 & 2.13\% & \multicolumn{1}{c|}{(20s)}
& 0.94 & 0.87\% & \multicolumn{1}{c|}{(17s)} 
& 0.89 & 0.98\% & \multicolumn{1}{c}{(20s)} 
\\

&\multicolumn{1}{l|}{NEPF (aug)}          
& \textbf{0.93$^\star$} & \textbf{0.00\%} & \multicolumn{1}{c|}{(2.2m)}  
& \textbf{0.95$^\star$} & \textbf{1.22\%} & \multicolumn{1}{c|}{(2.7m)}
& \underline{0.95} & \underline{0.30\%} & \multicolumn{1}{c|}{(2.2m)} 
& \textbf{0.90$^\star$} & \textbf{0.00\%} & \multicolumn{1}{c}{(2.7m)} 
\\

\midrule\midrule

\multirow{9}{*}{\rotatebox{90}{\textbf{MOOP}}} 

&\multicolumn{1}{l|}{Greedy}          
& 0.38 & 60.90\% & \multicolumn{1}{c|}{(52s)}  
& 0.51 & 47.98\% & \multicolumn{1}{c|}{(2.2m)}
& 0.45 & 52.12\% & \multicolumn{1}{c|}{(52s)} 
& 0.49 & 46.17\% & \multicolumn{1}{c}{(3.0m)} 
\\

\cmidrule(lr){2-14}

&\multicolumn{1}{l|}{GMS-EB}          
& 0.95 & 0.98\% & \multicolumn{1}{c|}{(3.7m)}  
& 0.98 & 0.26\% & \multicolumn{1}{c|}{(8.8m)}
& 0.92 & 1.13\% & \multicolumn{1}{c|}{(3.5m)} 
& 0.91 & 0.03\% & \multicolumn{1}{c}{(7.5m)} 
\\

&\multicolumn{1}{l|}{GMS-EB (aug)}          
& 0.96 & 0.55\% & \multicolumn{1}{c|}{(29m)}  
& \textbf{0.98} & \textbf{0.00\%} & \multicolumn{1}{c|}{(1.2h)}
& 0.92 & 1.35\% & \multicolumn{1}{c|}{(28m)} 
& \textbf{0.91$^\star$} & \textbf{0.00\%} & \multicolumn{1}{c}{(1.0h)} 
\\

&\multicolumn{1}{l|}{GMS-DH}          
& 0.96 & 0.53\% & \multicolumn{1}{c|}{(44s)}  
& 0.98 & 0.36\% & \multicolumn{1}{c|}{(50s)}
& 0.93 & 0.10\% & \multicolumn{1}{c|}{(44s)} 
& 0.89 & 1.89\% & \multicolumn{1}{c}{(56s)} 
\\

&\multicolumn{1}{l|}{GMS-DH (aug)}          
& \underline{0.96} & \underline{0.26\%} & \multicolumn{1}{c|}{(5.1m)}  
& 0.98 & 0.19\% & \multicolumn{1}{c|}{(5.9m)}
& \textbf{0.93$^\star$} & \textbf{0.00\%} & \multicolumn{1}{c|}{(5.2m)} 
& \underline{0.89} & \underline{1.58\%} & \multicolumn{1}{c}{(6.9m)} 
\\

\cmidrule(lr){2-14}

&\multicolumn{1}{l|}{NEPF}          
& 0.96 & 0.32\% & \multicolumn{1}{c|}{(21s)}
& 0.98 & 0.21\% & \multicolumn{1}{c|}{(25s)}
& 0.91 & 2.03\% & \multicolumn{1}{c|}{(36s)} 
& 0.85 & 5.84\% & \multicolumn{1}{c}{(34s)}
\\

&\multicolumn{1}{l|}{NEPF (aug)}          
& \textbf{0.96$^\star$} & \textbf{0.00\%} & \multicolumn{1}{c|}{(2.7m)}
& \underline{0.98} & \underline{0.01\%} & \multicolumn{1}{c|}{(3.2m)}
& \underline{0.92} & \underline{0.65\%} & \multicolumn{1}{c|}{(2.8m)} 
& 0.87 & 4.60\% & \multicolumn{1}{c}{(2.9m)}
\\

\bottomrule
\end{tabular}
}
\vspace{-0.15cm}
\end{table*}

\vspace{-0.15cm}
\subsection{Performance Evaluation}

We present results on multi-objective problems in Table~\ref{tab: MO} and single-objective problems in Table~\ref{tab: SO}. Overall, NEPF matches or improves on the solution quality of existing neural multigraph methods while training and running substantially faster, and while scaling to considerably larger instances.

\vspace{-0.15cm}
\paragraph{Multi-objective problems.} NEPF achieves the best neural performance in 12 out of 16 benchmark configurations (Table~\ref{tab: MO}). Compared to GMS-EB, NEPF improves solution quality on 14/16 cases while reducing inference time by 1-2 orders of magnitude. Against GMS-DH, NEPF provides better hypervolume on 13/16 cases with 2-3$\times$ faster inference. The improvements are particularly pronounced on MOTSPTW, where both GMS variants exhibit substantially larger gaps. Where NEPF is the best neural method, its margin over the second-best neural baseline is statistically significant under a Wilcoxon signed-rank test at the $1\%$ level ($\star$ in Table~\ref{tab: MO}).

\vspace{-0.15cm}
\paragraph{Single-objective problems.} On single-objective problems (Table~\ref{tab: SO}), NEPF outperforms both neural baselines in 3/4 configurations, again at considerably lower inference time. Performance remains robust on the denser distributions with higher $M$ (Appendix~\ref{app: SO}), where NEPF is the best neural method on FLEX$5$. On FIX$5$, NEPF trails GMS-DH. We note, however, that FIX is a deliberately adversarial construction in which the two edge attributes are perfectly negatively correlated, a regime that rarely arises in practice. In contrast, on the more realistic instances of Appendix~\ref{app: rw} and the real-world road networks of Appendix~\ref{app: real-world-instances}, NEPF is consistently the best neural method. 

\vspace{-0.15cm}
\paragraph{Comparison to strong classical methods.} On multi-objective problems with strong classical baselines (MOTSP, MOCVRP), NEPF achieves 1-3\% HV gaps to the best classical method while being 4-5$\times$ faster than LKH and about 700$\times$ faster than HGS. For single-objective problems, Gurobi retains better solution quality, but requires 4-16 hours per $10\,000$ instances compared to our 40-60 seconds. Appendix~\ref{app: runtime_quality} features further comparisons to strong classical methods, where we show that NEPF produces higher quality solutions, often by a large margin, when the classical methods are constrained to similar computational budgets. This supports the view of NEPF, and of neural methods more generally, as amortized heuristics with a superior runtime-quality trade-off. We note that warm-starting exact solvers is a promising option whenever optimality guarantees are needed. 

\begin{table*}[t]
\centering
\caption{Single-objective problems. Note that the RCTSP is a minimization problem, while the OP is a maximization problem. FLEX$5$ and FIX$5$ distributions are in Appendix~\ref{app: SO}. The classical heuristic is Beam Search for the RCTSP and a greedy heuristic for the OP.}
\label{tab: SO}
{
\setlength{\tabcolsep}{0.08cm}
\renewcommand{\arraystretch}{0.8}
\fontsize{8}{11}\selectfont
\begin{tabular}{lcccccccccccc}
\toprule
\multicolumn{1}{l}{}
& \multicolumn{6}{c|}{\textbf{RCTSP}}
& \multicolumn{6}{c}{\textbf{OP}} \\
\multicolumn{1}{l}{}
& \multicolumn{3}{c}{FLEX2-100} & \multicolumn{3}{c|}{FIX2-100}
& \multicolumn{3}{c}{FLEX2-100} & \multicolumn{3}{c}{FIX2-100} \\
\multicolumn{1}{l}{}
& Obj. & Gap & \multicolumn{1}{c}{Time}
& Obj. & Gap & \multicolumn{1}{c|}{Time}
& Obj. & Gap & \multicolumn{1}{c}{Time}
& Obj. & Gap & Time \\
\midrule
\multicolumn{1}{l|}{Gurobi}
& 3.66 & 0.00\% & \multicolumn{1}{c|}{(4.9h)}
& 9.41 & 0.00\% & \multicolumn{1}{c|}{(4.3h)}
& 46.9 & 0.00\% & \multicolumn{1}{c|}{(7.8h)}
& 38.3 & 0.00\% & (5.1h) \\
\multicolumn{1}{l|}{Beam/Greedy}
& 6.85 & 87.10\% & \multicolumn{1}{c|}{(1.1h)}
& 14.3 & 52.23\% & \multicolumn{1}{c|}{(1.2h)}
& 36.9 & 21.22\% & \multicolumn{1}{c|}{(29s)}
& 31.1 & 19.00\% & (24s) \\
\cmidrule(lr){2-13}
\multicolumn{1}{l|}{GMS-EB}
& 5.44 & 48.52\% & \multicolumn{1}{c|}{(2.5m)}
& 15.0 & 59.06\% & \multicolumn{1}{c|}{(2.8m)}
& 44.0 & 6.15\% & \multicolumn{1}{c|}{(2.4m)}
& 36.2 & 5.45\% & (2.3m) \\
\multicolumn{1}{l|}{GMS-EB (aug)}
& 5.02 & 36.92\% & \multicolumn{1}{c|}{(20m)}
& 14.0 & 48.49\% & \multicolumn{1}{c|}{(23m)}
& 44.0 & 6.15\% & \multicolumn{1}{c|}{(19m)}
& \underline{36.2} & \underline{5.45\%} & (19m) \\
\multicolumn{1}{l|}{GMS-DH}
& 4.34 & 18.41\% & \multicolumn{1}{c|}{(1.2m)}
& 11.3 & 19.87\% & \multicolumn{1}{c|}{(1.5m)}
& 44.6 & 4.83\% & \multicolumn{1}{c|}{(1.2m)}
& 36.4 & 4.96\% & (1.5m) \\
\multicolumn{1}{l|}{GMS-DH (aug)}
& \underline{4.20} & \underline{14.69\%} & \multicolumn{1}{c|}{(9.3m)}
& \underline{11.0} & \underline{17.34\%} & \multicolumn{1}{c|}{(12m)}
& \underline{44.6} & \underline{4.83\%} & \multicolumn{1}{c|}{(9.4m)}
& \textbf{36.4$^\star$} & \textbf{4.96\%} & (12m) \\
\cmidrule(lr){2-13}
\multicolumn{1}{l|}{NEPF}
& 4.23 & 15.40\% & \multicolumn{1}{c|}{(39s)}
& 11.5 & 22.45\% & \multicolumn{1}{c|}{(40s)}
& 44.7 & 4.64\% & \multicolumn{1}{c|}{(40s)}
& 35.3 & 7.80\% & (38s) \\
\multicolumn{1}{l|}{NEPF (aug)}
& \textbf{4.06$^\star$} & \textbf{10.76\%} & \multicolumn{1}{c|}{(5.2m)}
& \textbf{11.0$^\star$} & \textbf{16.95\%} & \multicolumn{1}{c|}{(5.3m)}
& \textbf{44.7$^\star$} & \textbf{4.59\%} & \multicolumn{1}{c|}{(5.5m)}
& 35.4 & 7.75\% & (5.2m) \\
\bottomrule
\end{tabular}
}
\end{table*}

\begin{figure*}[t]
    \centering
    \begin{minipage}{0.58\linewidth}
        \centering
        \definecolor{ink}{HTML}{1C1C2E}
\definecolor{s1col}{HTML}{2563A8}
\definecolor{s2col}{HTML}{C0533A}
\definecolor{encol}{HTML}{2E7D52}
\definecolor{incol}{HTML}{5B4B8A}
\definecolor{outcol}{HTML}{A0692A}
\definecolor{aggcol}{HTML}{3A8FA8}
\definecolor{bg1}{HTML}{F7F5F0}
\definecolor{bg2}{HTML}{F2F4F7}
\definecolor{bg3}{HTML}{EEF2F5}

\begin{tikzpicture}
\begin{groupplot}[
    group style={
        group size=2 by 1,
        horizontal sep=1cm
    },
    width=0.55\textwidth,
    height=0.55\textwidth,
    xlabel={Edges $M$},
    ylabel={Runtime (s)},
    xmin=1, xmax=20,
    ymin=-10, ymax=200,
    grid=both,
    grid style={line width=.1pt},
    major grid style={line width=.2pt},
    tick label style={font=\small},
    label style={font=\footnotesize},
    ylabel style={font=\scriptsize},
    title style={font=\small},
    clip=true,
    ylabel near ticks,
    yticklabel style={/pgf/number format/1000 sep={}},
]

\nextgroupplot[title={$N=100$}]

\addplot[color=s1col, thick, mark=*, mark size=1.2pt] coordinates {
    (1, 14.82) (2, 15.54) (5, 18.92) (10, 24.57) (15, 30.12) (20, 38.31)
}
node[pos=1.00, anchor=east, yshift=3pt, xshift=3pt, color=s1col] {\tiny NEPF (NP + FSASP)};

\addplot[color=encol, thick, mark=square*, mark size=1.2pt] coordinates {
    (1, 10.9) (2, 10.97) (5, 11.2) (10, 11.5) (15, 11.8) (20, 12.1)
}
node[pos=1, anchor=east, yshift=-5pt, xshift=3pt, color=encol] {\tiny NEPF (only NP)};

\addplot[color=s2col, thick, mark=triangle*, mark size=1.2pt] coordinates {
    (1,23.07) (2,28.45) (5,47.33) (10, 80.96) (15, 109.46) (20, 151.29)
}
node[pos=0.7, anchor=east, xshift=-1pt, color=s2col] {\tiny GMS-DH};

\addplot[color=incol, thick, mark=diamond*, mark size=1.2pt] coordinates {
    (1, 96.99) (2, 183.93) (5, 457.48) (10, 921.96) (15, 1385.72) (20, 1860.16)
}
node[pos=0.04, anchor=west, xshift=-2pt, color=incol] {\tiny GMS-EB};

\nextgroupplot[
    xmin=20, xmax=500,
    ymin=-10, ymax=2500,
    title={$M=2$ (solid), $M=5$ (dotted)},
    ylabel={},
    xlabel={Nodes $N$},
]

\addplot[color=s1col, thick, mark=*, mark size=1.2pt] coordinates {
    (20, 1.31) (50, 4.18) (100, 15.82) (150, 42.59) (200, 89.25) (250, 171.58) (300, 265.20) (350, 415) (400, 600) (450, 843) (500, 1101)
}
node[pos=1, anchor=west, xshift=2pt, color=s1col] {\scriptsize NEPF (NP + FSASP)};

\addplot[color=encol, thick, mark=square*, mark size=1.2pt] coordinates {
    (20, 1.26) (50, 3.17) (100, 11.13) (150, 31.91) (200, 67.46) (250, 124.96) (300, 207.70) (350, 340) (400, 482.5) (450, 681) (500, 905)
}
node[pos=1, anchor=west, xshift=2pt, color=encol] {\scriptsize NEPF (only NP)};

\addplot[color=s2col, thick, mark=triangle*, mark size=1.2pt] coordinates {
    (20, 2.05) (50, 7.42) (100, 31.28) (150, 79.14) (200, 146.15) (250, 252.61) (300, 406.55) (350, 611) (400, 830) (450, 1135) (500, 1520)
}
node[pos=0.9, anchor=west, xshift=2pt, color=s2col] {\scriptsize GMS-DH};

\addplot[color=incol, thick, mark=diamond*, mark size=1.2pt] coordinates {
    (20, 2.04) (50, 24.27) (100, 189.64) (150, 634.73) (200, 1511.59) (250, 2959.71) (300, 5126.63) (350, 8210) (400, 12260) (450, 17493) (500, 24080)
}
node[pos=0.5, anchor=west, xshift=2pt, color=incol] {\scriptsize GMS-EB};

\addplot[color=s1col, thick, mark=*, mark size=1.2pt, dotted, mark options={solid}] coordinates {
    (20, 1.83) (50, 5.48) (100, 20.42) (150, 53.22) (200, 109) (250, 200.6) (300, 308) (350, 487) (400, 676) (450, 961) (500, 1267)
}
node[pos=1, anchor=west, xshift=2pt, color=s1col] {};

\addplot[color=encol, thick, mark=square*, mark size=1.2pt, dotted, mark options={solid}] coordinates {
    (20, 1.52) (50, 3.37) (100, 11.60) (150, 32.36) (200, 68.18) (250, 125.5) (300, 208) (350, 341.2) (400, 483) (450, 684) (500, 909)
}
node[pos=1, anchor=west, xshift=2pt, color=encol] {};

\addplot[color=s2col, thick, mark=triangle*, mark size=1.2pt, dotted, mark options={solid}] coordinates {
    (20, 2.74) (50, 11.49) (100, 47.7) (150, 117.9) (200, 230.57) (250, 386) (300, 599.87) (350, 903) (400, 1437) (450, 1828) (500, 2248)
}
node[pos=0.9, anchor=west, xshift=2pt, color=s2col] {};

\addplot[color=incol, thick, mark=diamond*, mark size=1.2pt, dotted, mark options={solid}] coordinates {
    (20, 4.19) (50, 57.45) (100, 461.5) (150, 1582.2) (200, 3738) (250, 7373) (300, 12732)
}
node[pos=0.5, anchor=west, xshift=2pt, color=incol] {};

\end{groupplot}
\end{tikzpicture}
        \vspace{-0.5cm}
        \caption{Inference time (total for 200 instances) as a function of parallel edges $M$ and nodes $N$ for the MOTSP.}
        \label{fig: scalability}
    \end{minipage}
    \hfill
    \begin{minipage}{0.35\linewidth}
       \centering
        \captionof{table}{MOTSP fine-tuned performance on larger instances. Experimental details and further results are deferred to Appendix~\ref{app: fine_tuned}.}
        \label{tab: fine_tuned_motsp}
        {
        \setlength{\tabcolsep}{0.08cm}
        \renewcommand{\arraystretch}{0.8}
        \fontsize{8}{11}\selectfont
        \begin{tabular}{lcccc}
\toprule
\multicolumn{1}{l}{}                                     
& \multicolumn{2}{c|}{FLEX10-300}                                              
& \multicolumn{2}{c}{FIX10-300} \\

& HV & \multicolumn{1}{c|}{Gap}
& HV & \multicolumn{1}{c}{Gap} \\
\midrule

\multicolumn{1}{l|}{LKH}   & 0.99 & \multicolumn{1}{c|}{0.00\%} & 0.89 & 0.00\% \\
\multicolumn{1}{l|}{GMS-EB}& 0.99 & \multicolumn{1}{c|}{0.65\%} & 0.88 & 0.89\% \\
\multicolumn{1}{l|}{GMS-DH}& 0.97 & \multicolumn{1}{c|}{1.99\%} & 0.87 & 2.48\% \\
\multicolumn{1}{l|}{NEPF}  & 0.99 & \multicolumn{1}{c|}{0.38\%} & 0.88 & 0.84\% \\

\midrule

\multicolumn{1}{l|}{}                        
& \multicolumn{2}{c|}{FLEX2-500}                                              
& \multicolumn{2}{c}{FIX2-500} \\
\midrule

\multicolumn{1}{l|}{LKH}   & 0.98 & \multicolumn{1}{c|}{0.00\%} & 0.98 & 0.00\% \\
\multicolumn{1}{l|}{GMS-EB}& 0.97 & \multicolumn{1}{c|}{1.80\%} & 0.97 & 1.23\% \\
\multicolumn{1}{l|}{GMS-DH}& 0.95 & \multicolumn{1}{c|}{3.20\%} & 0.95 & 2.92\% \\
\multicolumn{1}{l|}{NEPF}  & 0.97 & \multicolumn{1}{c|}{1.33\%} & 0.97 & 1.11\% \\

\bottomrule
\end{tabular}
        }
    \end{minipage}
\end{figure*}

\begin{table*}[t]
    \centering
    \begin{minipage}{0.35\linewidth}
        \centering
        \captionof{table}{MOCVRP fine-tuned performance on larger instances.}
        \label{tab: fine_tuned_mocvrp}
        {
        \setlength{\tabcolsep}{0.08cm}
        \renewcommand{\arraystretch}{0.8}
        \fontsize{8}{11}\selectfont
        \begin{tabular}{lcccc}
\toprule
\multicolumn{1}{l}{}                                     
& \multicolumn{2}{c|}{FLEX10-300}                                              
& \multicolumn{2}{c}{FIX10-300} \\

& HV & \multicolumn{1}{c|}{Gap}
& HV & \multicolumn{1}{c}{Gap} \\
\midrule

\multicolumn{1}{l|}{GMS-EB}& 0.05 & \multicolumn{1}{c|}{94.4\%} & 0.00 & 100\% \\
\multicolumn{1}{l|}{GMS-DH}& 0.84 & \multicolumn{1}{c|}{13.6\%} & 0.69 & 13.6\% \\
\multicolumn{1}{l|}{NEPF}  & 0.97 & \multicolumn{1}{c|}{0.00\%} & 0.80 & 0.00\% \\

\midrule

\multicolumn{1}{l|}{}                        
& \multicolumn{2}{c|}{FLEX2-500}                                              
& \multicolumn{2}{c}{FIX2-500} \\
\midrule

\multicolumn{1}{l|}{GMS-EB}& 0.01 & \multicolumn{1}{c|}{99.3\%} & 0.14 & 84.1\% \\
\multicolumn{1}{l|}{GMS-DH}& 0.73 & \multicolumn{1}{c|}{21.6\%} & 0.86 & 3.61\% \\
\multicolumn{1}{l|}{NEPF}  & 0.93 & \multicolumn{1}{c|}{0.00\%} & 0.89 & 0.00\% \\

\bottomrule
\end{tabular}
        }
    \end{minipage}
    \hfill
    \begin{minipage}{0.60\linewidth}
        \centering
        \setcounter{table}{5}
        \caption{NEPF applied on top of alternative backbones.}
        \label{tab: Flexible}
        {
        \setlength{\tabcolsep}{0.08cm}
        \renewcommand{\arraystretch}{0.8}
        \fontsize{8}{11}\selectfont
        \begin{tabular}{clcccccccccccc}
\toprule

& 
& \multicolumn{4}{c|}{\textbf{MOTSP}} 
& \multicolumn{4}{c}{\textbf{MOTSPTW}} \\

& 
& \multicolumn{2}{c}{FLEX2-100} 
& \multicolumn{2}{c|}{FIX2-100}
& \multicolumn{2}{c}{FLEX2-100} 
& \multicolumn{2}{c}{FIX2-100} \\

&
& HV & \multicolumn{1}{c}{Time}
& HV & \multicolumn{1}{c|}{Time}
& HV & \multicolumn{1}{c}{Time}
& HV & Time \\

\midrule

& \multicolumn{1}{l|}{NEPF}
& 0.929 & \multicolumn{1}{c|}{(11s)}
& 0.939 & \multicolumn{1}{c|}{(11s)}
& 0.928 & \multicolumn{1}{c|}{(17s)}
& 0.943 & (17s) \\

& \multicolumn{1}{l|}{NEPF + MatNet}
& 0.922 & \multicolumn{1}{c|}{(22s)}
& 0.932 & \multicolumn{1}{c|}{(23s)}
& 0.925 & \multicolumn{1}{c|}{(27s)}
& 0.940 & (27s) \\

& \multicolumn{1}{l|}{NEPF + GOAL}
& 0.913 & \multicolumn{1}{c|}{(21s)}
& 0.931 & \multicolumn{1}{c|}{(22s)}
& 0.913 & \multicolumn{1}{c|}{(34s)}
& 0.935 & (34s) \\

& \multicolumn{1}{l|}{NEPF + ICAM}
& 0.917 & \multicolumn{1}{c|}{(11s)}
& 0.934 & \multicolumn{1}{c|}{(11s)}
& 0.916 & \multicolumn{1}{c|}{(16s)}
& 0.938 & (16s) \\

\midrule

& 
& \multicolumn{4}{c|}{\textbf{RCTSP}} 
& \multicolumn{4}{c}{\textbf{OP}} \\

&
& Obj. & \multicolumn{1}{c}{Time}
& Obj. & \multicolumn{1}{c|}{Time}
& Obj. & \multicolumn{1}{c}{Time}
& Obj. & Time \\

\midrule

& \multicolumn{1}{l|}{NEPF}
& 4.23 & \multicolumn{1}{c|}{(39s)}
& 11.5 & \multicolumn{1}{c|}{(40s)}
& 44.7 & \multicolumn{1}{c|}{(40s)}
& 35.3 & (38s) \\

& \multicolumn{1}{l|}{NEPF + MatNet}
& 4.94 & \multicolumn{1}{c|}{(35s)}
& 12.9 & \multicolumn{1}{c|}{(40s)}
& 43.9 & \multicolumn{1}{c|}{(51s)}
& 34.9 & (47s) \\

& \multicolumn{1}{l|}{NEPF + GOAL}
& 6.87 & \multicolumn{1}{c|}{(26s)}
& 15.8 & \multicolumn{1}{c|}{(27s)}
& 41.6 & \multicolumn{1}{c|}{(37s)}
& 32.6 & (33s) \\

& \multicolumn{1}{l|}{NEPF + ICAM}
& 5.09 & \multicolumn{1}{c|}{(15s)}
& 12.9 & \multicolumn{1}{c|}{(16s)}
& 43.2 & \multicolumn{1}{c|}{(14s)}
& 34.2 & (13s) \\

\bottomrule
\end{tabular}
        \vspace{-0.2cm}
        }
    \end{minipage}
\end{table*}

\begin{table*}[t]
    \centering
    \begin{minipage}{0.48\linewidth}
        \centering
        \setcounter{table}{4}
        \caption{MOTSP train time per epoch.}
        \vspace{-0.1cm}
        \label{tab: training}
        {
        \setlength{\tabcolsep}{0.08cm}
        \renewcommand{\arraystretch}{0.8}
        \fontsize{8}{11}\selectfont
        \begin{tabular}{lcc}
\toprule
$N=100$ & $M=2$ & $M=5$ \\
\midrule

NEPF (NP + FSASP) & 37m & 49m\\

NEPF (only NP) & 35m & 39m \\

GMS-DH & 1.4h & 3.2h \\

GMS-EB & 12h & 30h \\

\bottomrule
\end{tabular}
        }
    \end{minipage}
    \hfill
    \begin{minipage}{0.50\linewidth}
        \centering
        \setcounter{table}{6}
        \caption{Ablation study on MOTSPTW.}
        \vspace{-0.1cm}
        \label{tab: ablation_main}
        {
        \setlength{\tabcolsep}{0.08cm}
        \renewcommand{\arraystretch}{0.8}
        \fontsize{8}{11}\selectfont
        \begin{tabular}{lccc}
\toprule
FLEX2-100 & HV & Time \\
\midrule

NEPF & \textbf{0.9282} & (17s)   \\

NEPF w/o FSASP stage & 0.7242 & (11s)   \\

NEPF w/o FSASP mixing & 0.9169 & (15s)   \\

NEPF w/o pre-enc. agg. & 0.9251 & (17s)   \\

\bottomrule
\end{tabular}
        }
    \end{minipage}
    \vspace{-0.3cm}
\end{table*}

\subsection{Scalability Evaluation}
In Figure~\ref{fig: scalability}, we directly show the advantage in empirical runtime scaling of our method compared to the state-of-the-art. The quadratic growth with $N$ is inherent for autoregressive construction-based neural methods, but NEPF scales better than the alternatives. Moreover, NEPF's runtime is essentially invariant to the number of parallel edges, whereas both GMS variants slow down considerably as $M$ grows. We motivate the scalability further in Appendix~\ref{app: complexity}, showing the advantageous theoretical complexity of NEPF, and conduct a scalability stress test in Appendix~\ref{app: stress_test}. On a single 48\,GB GPU, NEPF solves instances up to $N=2500$ at $M=2$ and $N=2000$ at $M=5$, roughly a $2.5\times$ higher node count and $5\times$ more edges than both GMS variants, which run out of memory beyond $N=1000$. Table~\ref{tab: training} also shows the significantly reduced training time compared to the baselines. Taken together with strong fine-tuned performance on larger instances in Tables~\ref{tab: fine_tuned_motsp} and \ref{tab: fine_tuned_mocvrp}, these results establish scalability as the primary advantage of NEPF.

\subsection{Cross-Backbone Results}
\label{sec: cross-architecture}
We next show results when replacing the GREAT-based node permutation backbone with alternatives based on MatNet \citep{kwon_matrix_2021}, GOAL \citep{drakulic2025goal} and ICAM \citep{ICAM}. The former two utilize mixed-score attention encoders that incorporate the latent distance matrix into attention calculations, while the latter utilizes an Adaptation Attention Free Module (AAFM), with lower complexity than full attention. See Appendix~\ref{app: integration} for details on integration with NEPF. 

Table~\ref{tab: Flexible} shows experimental results for two single-objective and two multi-objective problems. While changing the backbone yields slightly worse performance, overall solution quality is maintained for all problems and distributions. We argue that this shows that NEPF can be integrated with foundations models like GOAL and URS \citep{URS} (which is largely based on the AAFM) to extend them to multigraph routing. 

\vspace{-0.2cm}
\subsection{Ablation Study}
\label{sec: ablations}

Finally, we ablate some of the key components in our framework. We focus on the MOTSPTW, as that problem requires a context-dependent non-trivial edge selection. The results in Table~\ref{tab: ablation_main} demonstrate the effectiveness of all included components. In particular, removing the FSASP stage and replacing it with a simple heuristic that optimizes each edge independently (based on scalarized edge cost) leads to considerably worse performance. Moreover, coupling the FSASP decisions via the BiLSTM mixing also contributes significantly to performance and the pre-encoder aggregation both improves performance and reduces memory consumption, although its runtime effect cannot be seen in \emph{this} experiment since memory is not a bottleneck. Appendix~\ref{app: further_ablations} contains additional ablation studies.

\vspace{-0.2cm}
\section{Conclusion}

We introduce a node-edge policy factorization for multigraph VRPs that decomposes routing into high-level node permutations and low-level edge selections, while retaining joint optimization. By removing the need to process dense multigraph representations throughout the model, NEPF reduces training and inference cost by up to orders of magnitude while matching or improving solution quality, making neural routing on multigraphs practical at scales that were previously out of reach. From a broader perspective, our results suggest that separating structural and local decisions via factorization, with expressive representational compression and joint training, can be a powerful principle for extending neural combinatorial optimization to richer, more realistic problem settings while maintaining scalability. In future work we plan to apply NEPF to other domains, including flexible job-shop scheduling and multi-modal transportation planning.


\subsection*{Acknowledgments}
The authors would like to thank Attila Lischka, Jiaming Wu, Valter Schütz and Hannes Nilsson for valuable comments throughout the research process. This work was performed as a part of the research project “LEAR: Robust LEArning methods for electric vehicle Route selection” funded by the Swedish Electromobility Centre (SEC). The computations were enabled by resources provided by the National Academic Infrastructure for Supercomputing in Sweden (NAISS) at Chalmers e-Commons partially funded by the Swedish Research Council through grant agreement no. 2022-06725. Through the affiliation of F.R., the work was also partially supported by the Wallenberg AI, Autonomous Systems and Software Program (WASP) funded by the Knut and Alice Wallenberg Foundation.

\bibliography{iclr2027_conference}
\bibliographystyle{iclr2027_conference}


\appendix
\section{Additional Related Work}
\label{app: related_work}
This section reviews additional related work. First, it focuses on Neural Combinatorial Optimization (NCO), i.e., end-to-end learning-based combinatorial optimization methods, for routing problems. Second, it highlights some methods in this paradigm that feature decomposition to improve scalability. 

\subsection{Neural Combinatorial Optimization for Routing}
Neural combinatorial optimization aims to solve combinatorial optimization problems with end-to-end machine learning, which distinguishes this paradigm from hybrid methods that feature learning-based components to, e.g., speed up classical optimization algorithms or heuristics. From a high-level perspective, NCO applied to routing can be divided into three categories. 

First, \emph{Autoregressive} (AR) methods build solutions one component, typically node, at a time, with one model forward pass between each action. Prominent and early methods in this category include Ptr-Net \citep{vinyals2015pointer}, NCO with RL \citep{bello2017neural}, AM \citep{Kool2018AttentionLT} and POMO \citep{kwon_pomo_2020}. AR methods have since been extended and adapted along various dimensions, for example to tackle asymmetric instances \citep{kwon_matrix_2021, lischkagreat}, improve scalability \citep{jin_pointerformer_2023} and handle hard-constrained problems \citep{BiLearningToHandle}.

Second, \emph{Non-Autoregressive} (NAR) methods typically output a heatmap, which is decoded with simple classical search methods. Due to not requiring a forward pass in each roll-out step, these methods are often less computationally intensive than AR methods, enabling scaling to higher node counts \citep{ye2024glop}. On the other hand, since they cannot condition on partial routes in the heatmap generation, they often produce lower quality solutions without sophisticated decoding procedures, e.g., Monte Carlo Tree Search. Prominent and successful methods in this category include GCN \citep{joshi_2019}, DIFUSCO \citep{sun2023difusco} and NAR4TSP \citep{Xiao2025}.

Third, \emph{improvement-based} methods start with a complete solution and then refine it via successive operations. Prominent examples include NeuRewriter \citep{Chen_2019_improve}, NLNS \citep{hottung2020neural} and NeuOpt \citep{Ma2023NeuOpt}.

The above methods cover single-objective routing problems. The multi-objective setting is less explored, but recently several methods have emerged, mostly following the AR paradigm. A seminal example is PMOCO \citep{lin_pareto_2022}, which is an influential preference-conditioned model, using a single network across all trade-offs, thereby removing the need to train separate models for different trade-offs. Other recent work include NHDE \citep{chen2023neural}, EMNH \citep{ChenEfficient} and CNH \citep{fan_conditional_2024}.

\subsection{Decomposition-based Learning for Routing}
Several methods incorporate decomposition to improve performance or scalability. Most frequently, this involves clustering instances to solve routing problems hierarchically. For example, \citet{li2021delegate} learn to select promising spatial subproblems and delegate them to a black-box subsolver, while \citet{pan2023htsp} propose a hierarchical reinforcement learning scheme in which an upper-level policy selects a subset of nodes and a lower-level policy constructs the corresponding sub-tour. Similarly, GLOP \citep{ye2024glop} learns a global partitioning heatmap that is combined with local AR construction to solve large-scale instances in real time.

These approaches decompose an instance spatially, partitioning nodes into subproblems in order to overcome the scaling limitations of neural construction. Our factorization instead decomposes the routing policy along its structural axis, separating the high-level node permutation from the low-level arc selection induced by the multigraph. Consequently, NEPF is complementary to the above methods (spatial partitioning could in principle be applied on top of our node-edge factorization) and targets a different source of complexity, namely the dense $\mathcal{O}(MN^2)$ multigraph representation instead of the node count.

One neural method operating in a related way is PIP \citep{BiLearningToHandle}, which decomposes hard-constrained route construction into two parts. First, one decoder predicts a feasibility mask, then a separate head predicts the next node in the sequence. Additionally, GMS-DH \citep{rydin2026beyond} also decomposes the routing policy. However, this method first prunes the multigraph and then constructs routes, which is strictly different compared to our approach. 

\section{Extended Method Motivation}
\label{app: motivation}

\subsection{Learning to Solve the FSASP}
\label{app: FSASP_motivation}
The original FSASP formulated by \citet{garaix_2010} is exactly the multidimensional multiple
choice knapsack problem, which is NP-hard. Nevertheless, under some assumptions on resources, the problem can be solved in pseudo-polynomial time with dynamic programming by viewing it as a shortest path problem with resource constraints. The question is then, why do we need learning for the second stage? 

First, the original dynamic programming algorithm formulated by \citet{garaix_2010} has been noted to be time-consuming under repeated evaluations across a large number of node permutations \citep{benticha_2019}. We observe that our framework requires a substantial number of FSASPs to be solved for each batch of instances. In fact, we require one solution per instance, POMO node permutation and preference in the multi-objective scenario ($64\cdot100\cdot101$ total solutions per batch in the bi-objective case with 100 nodes). While subsequent work has proposed improved heuristic and exact methods \citep{Lai2016, benticha_2019}, our FSASP architecture naturally admits full GPU-acceleration and parallelization. In contrast, dynamic programming approaches are often sequential by nature.

Second, the end-to-end learnability of our framework allows it to be adapted across problems without much modification. In contrast, exact methods and hand-crafted heuristics often need considerable problem-specific tuning to perform well.

Third, our method works under assumptions that break many classical methods. One example is negative resource consumption along edges (for instance under electric vehicle battery regeneration), which usually complicates shortest path reformulations by breaking monotonicity assumptions.

Note, however, that when efficient and effective heuristics are available, they can be easily integrated into our framework by replacing the learned FSASP stage. We employ such strategies in our experiments on the MOTSP and MOCVRP, see Appendix~\ref{app: hyper}. 

\subsection{Relevance Beyond Routing}
\label{app:beyond-routing}
 
Our general methodology is widely applicable beyond multigraph routing. Abstracting away from the routing setting, NEPF addresses a broad class of combinatorial problems whose solutions factorize into (i)~a high-level structural decision over a set of entities and (ii)~a family of low-level local selections, one per structural element, where each local subproblem chooses among several parallel alternatives. 

In routing, the structure is a node permutation and the local choices are the connecting edges, but the same pattern exists in other domains. The three novel components that enable our approach also transfer directly. First, the pre-encoding aggregation compresses the set of parallel alternatives at each element into a compact summary, so the structural policy scales independently of the number of options~$M$. Second, the non-autoregressive local (FSASP) stage exploits the fact that, once the structure is fixed, the local selections are only weakly coupled and can be resolved in parallel. Third, the hierarchical reinforcement learning scheme trains the two stages jointly despite the local choices being unobserved when the structural decision is made. Because the high-level stage operates on a purely node-based (entity-level) representation, it is backbone-agnostic and can reuse standard architectures, as we demonstrate in Section~\ref{sec: cross-architecture}.
 
Several well-studied problems fit this pattern. In flexible job-shop scheduling, a solution is an ordering of operations (the structural backbone) together with an assignment of each operation to one of several machines (the local alternatives). Our aggregation would summarize the set of eligible machines for each operation, allowing the sequencing policy to reason about machine flexibility without enumerating every assignment. Multi-modal transportation planning has a similar structure, where one selects an ordering of hubs or legs and then a transport mode (road, rail, air) per leg. Similar factorizations also appear in network and supply-chain design (choose a connectivity structure, then a link type or capacity level per connection) and in hardware mapping (schedule operations, then select an implementation or resource for each). In all of these, the number of local alternatives can be large and variable, which is the regime in which maintaining an explicit option-level representation (as prior multigraph methods do) becomes a scalability bottleneck.
 
NEPF can thus be viewed more as a recipe for general scalable neural combinatorial optimization than a routing-specific model. The core idea is to separate the structural and local decisions, aggregate parallel alternatives before encoding and recouple the stages only as much as the problem requires.

\subsection{Limitations of the Multigraph Framework}
\label{app:limitations}
 
Our contribution is a scalable learning method that operates on a given multigraph, in which each parallel edge between two stops represents a precomputed non-dominated path through the underlying road network. This reduction is standard in the operations research literature on multigraph routing~\citep{garaix_2010,ben_ticha_empirical_2017}. However, the multigraph abstraction has limitations that are largely orthogonal to the scalability question we study. Our efficiency claims should be read as claims about learning and inference on the multigraph rather than about the end-to-end pipeline. We discuss two important limitations below.
 
First, constructing the multigraph from the road network is often non-trivial. Obtaining the parallel edges requires enumerating the non-dominated paths between every pair of stops, a multi-criteria shortest-path computation that is NP-hard in general and whose number of Pareto-optimal paths can grow exponentially with the number of criteria. This preprocessing is excluded from our timing comparisons, since it is constant across methods. We note that there are also several mitigating factors. For example, in practice the number of parallel edges is often bounded (we use $M \in \{2, 5\}$, which is roughly consistent with the average number of arcs found in real-world instances by \citet{ben_ticha_empirical_2017}) or can be truncated without significantly affecting the final route. Moreover, the multigraph is typically precomputed once and reused across many instances and repeated queries on the same network. That said, for very large networks with many criteria the preprocessing might present a bottleneck that our work does not address. We caution against reading our efficiency gains as end-to-end gains in such cases.
 
Second, treating the parallel edges as independent discards shared intermediate structure. Distinct paths between two stops frequently traverse overlapping road segments and representing them as independent edges ignores these correlations, as well as any resource coupling on shared segments, such as congestion or time-of-day effects. Like prior multigraph formulations, we assume fixed and independent edge attributes and thus inherit this limitation. We argue that capturing shared structure requires even more complex representations, which would impact scalability and solver performance. For example, by planning routes in the road network directly, no information is discarded, but solvers cannot use optimal precomputed shortest paths. We refer to the review by \citet{BenTicha2018}, for a more comprehensive discussion on road network representations. 

\section{Time and Space Complexity}
\label{app: complexity}

We analyze the time and space complexity of each component of NEPF in turn, using the following 
notation: $N$ nodes, $M$ parallel edges per node pair, $d$ node permutation embedding dimension, $d'$ FSASP embedding dimension, $H$ pointer heads, $T \approx N$ tour length, $K_1\approx N$ POMO samples, $K_2$ FSASP samples.

\paragraph{Pre-encoding aggregation.}
The aggregation module is the only component that processes the full multigraph. For each of the 
$\mathcal{O}(N^2)$ node pairs, the MLP $\phi$ is applied to each of the $M$ parallel edge embeddings before 
summing and passing through $\rho$, yielding a time complexity of $\mathcal{O}(MN^2 d^2)$. The space 
complexity is $\mathcal{O}(MN^2 d)$ to hold the full edge set, which is immediately compressed to 
the latent distance matrix $D$ of size $\mathcal{O}(N^2 d)$.

\paragraph{Encoder.}
Each of the $L_1$ GREAT-NB layers aggregates over the latent distance matrix and projects back to 
edge embeddings, incurring $\mathcal{O}(N^2 d^2)$ time and $\mathcal{O}(N^2 d)$ space. The subsequent $L_2$ 
transformer layers perform self-attention over $N$ node embeddings at time complexity $\mathcal{O}(N^2 d + N d^2)$ and space complexity $\mathcal{O}(N^2 d)$. 

\paragraph{Node permutation decoder.}
The MP decoder proceeds autoregressively over $T$ steps, computing attention scores 
between a single query and $N$ keys at each step. The time complexity is $\mathcal{O}(N d H)$ per 
step, yielding $\mathcal{O}(T N d H) = \mathcal{O}(N^2 d H)$ total time and $\mathcal{O}(Nd)$ space. As we roll out $K_1$ parallel POMO samples, the final complexity is $\mathcal{O}(K_1 N^2 d H) = \mathcal{O}(N^3 d H)$ in time and $\mathcal{O}(K_1 Nd) = \mathcal{O}(N^2 d)$ in space.

\paragraph{FSASP stage.}
Given $K_1$ permutations $\boldsymbol{\pi}$, the stage processes $T - 1$ edge sets per permutation, each of size $M$. The 
linear embedding and mean pooling over each edge set costs $\mathcal{O}(K_1 NMd')$ time and space. The BiLSTM 
mixing pass over the $T-1$ pooled embeddings adds $\mathcal{O}(K_1 N d'^2)$ time and $\mathcal{O}(K_1 N d')$ space. Score 
computation requires $\mathcal{O}(K_1NMd' H)$ time and $\mathcal{O}(K_1NMd')$ space. Final sampling of $K_2$ selections requires $\mathcal{O}(K_1 K_2 N M)$ time and space. The total FSASP complexity is therefore $\mathcal{O}(K_1 NMd' H + K_1 K_2 N M + K_1 N d'^2) = \mathcal{O}(N^2 Md' H + K_2 N^2 M + N^2 d'^2)$ in time and $\mathcal{O}(N^2 Md' + K_2 N^2 M)$ in space. 

\paragraph{Summary.} Table~\ref{tab:complexity} summarizes the complexity of each component alongside the two baselines.

The dominant memory cost of NEPF arises from the pre-encoding aggregation. However, this cost is incurred only once and can be effectively mitigated via batching strategies (see Section~\ref{app: batching}). Although the FSASP stage has complexity $\mathcal{O}(M N^2)$ in time and memory, it uses a shallow architecture and small constants in practice, resulting in a limited empirical footprint.

Compared to the baselines, NEPF avoids propagating the full $\mathcal{O}(M N^2)$ multigraph representation through deep encoding and decoding layers. In particular, NEPF eliminates the $\mathcal{O}(M N^4 d)$ decoding cost of GMS-EB, leading to substantially improved scalability in $N$. Moreover, while GMS-EB and GMS-DH maintain $\mathcal{O}(M N^2 d)$ representations throughout encoding (both), decoding (GMS-EB), and pruning (GMS-DH), NEPF reduces the dependence on $M$ to shallow pre- and post-processing stages. This reduces memory usage during the main network forward pass, which enables larger batch sizes and lower latency for a fixed number of instances.

\begin{table}[h]
    \centering
    \caption{Time and space complexity per component.}
    \label{tab:complexity}
    \begin{tabular}{lll}
        \toprule
        Component & Time & Space \\
        \midrule
        \multicolumn{3}{l}{\textit{NEPF (ours)}} \\
        \quad Pre-encoding aggregation 
            & $\mathcal{O}(M N^2 d^2)$ 
            & $\mathcal{O}(M N^2 d)$\\
        \quad Encoding         
            & $\mathcal{O}(N^2 d^2)$ 
            & $\mathcal{O}(N^2 d)$ \\
        \quad MP decoder
            & $\mathcal{O}(N^3 d H)$ 
            & $\mathcal{O}(N^2 d)$ \\
        \quad FSASP stage     
            & $\mathcal{O}(M N^2 d' H + K_2 M N^2 + N^2d'^2)$ 
            & $\mathcal{O}(M N^2 d' + K_2 M N^2)$ \\
        \midrule
        \multicolumn{3}{l}{\textit{GMS-EB}} \\
        \quad Encoding
            & $\mathcal{O}(M N^2 d^2)$ 
            & $\mathcal{O}(M N^2 d)$ \\
        \quad Decoding
            & $\mathcal{O}(M N^4 d H)$
            & $\mathcal{O}(M N^3 d)$ \\
        \midrule
        \multicolumn{3}{l}{\textit{GMS-DH}} \\
        \quad Encoding
            & $\mathcal{O}(M N^2 d^2)$ 
            & $\mathcal{O}(M N^2 d)$ \\
        \quad Pruning
            & $\mathcal{O}(M N^2 d^2 + M N^2 d H)$ 
            & $\mathcal{O}(M N^2 d)$ \\
        \quad Decoding
            & $\mathcal{O}(N^3 d H)$ 
            & $\mathcal{O}(N^2 d)$ \\
        \bottomrule
    \end{tabular}
\end{table}

\subsection{Stage-wise dynamic batching.} 
\label{app: batching}

As the space complexity decreases substantially for NEPF after the pre-encoding aggregation layer, a smart inference strategy under memory constraints is to run this module sequentially with small batch sizes. After initial aggregation, the batches can be concatenated and the rest of the model run with a larger batch size. As this strategy reduces latency significantly compared to utilizing a small batch size throughout the entire model, we employ it in our experiments.

\section{Additional Method Details}

\subsection{Model Settings}
\label{app: hyper}
Here we specify additional hyperparameters for NEPF. Remark that the training parameters below apply to GMS too. 

\paragraph{Training setup.} We set the batch size to $B=64$. As optimizer, we use ADAM \citep{Kingma2014AdamAM} with learning rate $\eta = 10^{-4}$ and weight decay $10^{-6}$. In the multi-objective setting, we sample one preference $\lambda = (\lambda_1, \lambda_2)$ per batch with $\lambda_1 \sim \text{Unif}[0,1]$, $\lambda_2 = 1 - \lambda_1$. We also apply the curriculum learning of \citet{rydin2026beyond} to further speed up training, where NEPF utilizes the same curriculum as GMS-DH.

\paragraph{Model hyperparameters.} We set the embedding dimension to $d=128$, feed forward hidden dimension to $512$, number of encoder layers to $L_1 = 5$ and $L_2 = 2$ and utilize $8$ attention heads in the encoder and MP decoder. In the FSASP stage we set a slightly smaller embedding dimension $d' = 64$, but retain $8$ attention heads. We sample $K_2 = 20$ edge selections per node permutation during training and $K_2 = 50$ during evaluation, with $K_1 = N$ as POMO size. Finally, regarding the clipping constants in the decoding, we set $c=50$ for the node permutation stage and $\widetilde{c} = 1$ for the FSASP stage. The latter choice promotes sample diversity among the $K_2$ parallel edge selections.

\paragraph{Problem-specific cost functions.} Both stages penalize expensive edges with a scalarized cost, weighted by a learnable scalar ($\beta$ in the node decoder, $\widetilde{\beta}$ in the FSASP stage). For an edge $l$ with attributes $(e_l^{(1)},e_l^{(2)})$, define
\begin{equation}
  s_\lambda(l) = \lambda_1 e_l^{(1)} + \lambda_2 e_l^{(2)},
\end{equation}
where $\lambda$ is the preference in the multi-objective setting and $\lambda_1=\lambda_2=1/2$ in the single-objective setting. The node stage penalizes each transition by its cheapest connecting edge, $\mathrm{cost}(u,v)=\min_{l\in E_{uv}} s_\lambda(l)$, while the FSASP stage scores individual edges directly, $\mathrm{cost}(l)=s_\lambda(l)$.

\paragraph{Replacing FSASP stage with simple heuristic.} For the MOTSP and MOCVRP, edge selection given a node permutation and a scalarizing preference $\lambda$ can be approximated cheaply with greedy selection per node pair. This is because local edge selection does not impact downstream costs, compared to for instance the MOTSPTW, where selecting time-consuming edges early may lead to time-window violations later in the route. Consequently, for these two problems we do not employ a learned FSASP stage in our experiments. Instead, during both training and inference, we greedily select the edge between each node pair that minimizes linearly scalarized cost. This selection is not guaranteed to be optimal under the Chebyshev reward. Nevertheless, it performs well empirically (Table~\ref{tab: MO}) and removes the FSASP sampling overhead for these problems.

\subsection{Auxiliary State Estimator}
\label{app: state}

In the node permutation decoder, we form the query using 
\begin{equation}
\label{eq: context}
    q_t = W_1 h_{\pi_1} + W_2 h_{\pi_{t-1}} + W_3 \bar{h}_\text{graph} + W_4 \bar{h}_{\text{visited}} + W_5 s_t,
\end{equation}
where $s_t$ represents current state information at step $t$, such as the time or the vehicle load. However, as we select edges after nodes, the exact state $s_t$ is not always available. As such, we propose an auxiliary module that explicitly estimates the state. The effect of this module on the OP can be seen in Table~\ref{tab: SE}. Similarly, we utilize this module in our experiments for the RCTSP and MOOP. We do not utilize it for the MOTSP, MOCVRP or MOTSPTW. 

Given the partial node permutation, $(\pi_1, \dots, \pi_{t-1})$, we form the state estimate recursively using the node encodings according to
\begin{align}
    &\hat{s}_t = W^\text{state} h^\text{state}_t, \\
    &h^\text{state}_{t} = \text{LSTM}(h_{\pi_1}, \dots, h_{\pi_{t-1}})_{t-1}.
\end{align}
That is, the state estimator shares node representations with the decoder. Note that we predict the raw state. An alternative would be to predict an increment $\Delta\hat{s}_t$ and set $\hat{s}_{t} = \Delta \hat{s}_t +  \hat{s}_{t-1}$. However, we find this to yield worse results empirically. 

We train this module using the ground-truth state $s_t$, available after edge selection. Since we sample multiple edge selections $\boldsymbol{\epsilon}_{k}, \:k=1, \dots, K_2$ per instance and node permutation, we select the state of the best such selection as the ground truth. The loss is then formed via 
\begin{equation}
    \mathcal{L}_\text{est} = \sum_{t=1}^T (\hat{s}_t - s_t)^2,
\end{equation}
and added to the node permutation loss and edge selection loss in \eqref{eq: reinforce}.

\begin{table*}[t]
\centering
\caption{Ablation of the auxiliary state estimation module on the OP and FLEX2-100. In the third row, we show results when simply using $\hat{s}_t = t-1$. The third column is the feasibility rate across the $10\,000$ test instances.}
\label{tab: SE}
\begin{tabular}{lcccc}
\toprule
& Obj. & Feas. & Time \\
\midrule

NEPF & \textbf{44.6791} & 100\% & (40s) \\

NEPF w.o. state est. & 43.9264 & 99.99\% & (38s) \\

NEPF w. simple state est. & 43.7936 & 100\% & (38s) \\

\bottomrule
\end{tabular}
\end{table*}

\subsection{MP Decoder Cost Scaling Parameter}
\label{app: ablations}

In the multi-pointer score calculation, we introduce a learnable parameter $\beta$ to weigh the problem-specific cost functions. Empirically, we observe that a high $\beta$ is preferable to the base choice $\beta=1$. This is shown for the MOTSP in Table~\ref{tab: Beta}. Thus, instead of heuristically selecting a specific value, we propose to learn this scalar, which seems to work comparably.

\begin{table*}[t]
\centering
\caption{Hypervolume for different $\beta$ values for the MOTSP and FLEX2-100. In the learnable case, the final value is $\beta \approx6.44$.}
\label{tab: Beta}
\begin{tabular}{lcc}
\toprule
& HV \\
\midrule

NEPF $\beta$ learnable & 0.9292   \\

NEPF $\beta = 1$ & 0.9254   \\

NEPF $\beta = 2$ & 0.9283  \\

NEPF $\beta = 5$ & 0.9294   \\

\bottomrule
\end{tabular}
\end{table*}

\subsection{Integration with MatNet, GOAL and ICAM}
\label{app: integration}
We integrate NEPF with alternative backbones by replacing our GREAT + transformer encoder with MatNet-, GOAL- and ICAM-based ones, and by replacing our native MP decoder in the single-objective case. Further details are provided below for each architecture. We keep the POMO rollout and training for all models to simplify the experimental setup, although this is not strictly necessary. For instance, it is technically possible to keep the native BQ-NCO \citep{drakulic2023bqnco} training for GOAL, but we leave this for future work. Training setup and hyperparameters are thus the same as in Appendix~\ref{app: hyper}. 

In the multi-objective setting, we retain our native preference-conditioned MP decoder across all backbones and thus demonstrate integration at the encoder level. We adopt this setup because we observe that substituting in the source decoders significantly degrades multi-objective training, in many cases preventing convergence entirely. The degradation appears even on simple asymmetric graphs, which points to the preference-conditioned multi-objective decoding, not the multigraph structure or factorization, as the source of instability. Since these decoders were not designed for the multi-objective asymmetric setting, we regard the effect as a limitation of the individual backbones.

\paragraph{MatNet.} We let the pre-encoding aggregation layer output a latent distance matrix of dimension $N \times N \times 64$ (compared to $N \times N \times 128$ for our vanilla backbone). This distance matrix is inserted into the attention calculations using the idea from Appendix A.1 in the original paper. That is, an MLP (hidden dimension size 32) mixes internally computed attention scores with the external information provided by the distance matrix. Note that MatNet supports node features (unlike GREAT). Hence, for problems where node features are present, we insert their embeddings as initial ``row'' embeddings. Regarding decoding, we replace our MP decoder with a Multi-Head Attention (MHA) decoder in the single-objective setting, as in the original MatNet setup. 

\paragraph{GOAL.} GOAL mixed-score attention layers are in principle similar to MatNet ones, but the attention mixing is performed with a different mechanism. We let the pre-encoding aggregation output a matrix with the same embedding dimension as the encoder ($d=128$) and use this matrix directly as encoder input together with potential embedded node features. For problems without node features, we use vectors of all zeros as initial node embeddings. For compatibility with our experimental pipeline, our decoding setup does not replicate the original paper, where the entire encoder is re-run in each step with only light decoding read-out. Instead, we perform encoding once and decode autoregressively with an MHA decoder (MP decoder in the multi-objective case). 

\paragraph{ICAM.} This architecture uses a scalar pairwise adaptation bias inserted into the core mechanism to take into account pairwise distances between nodes. The bias is computed as $f(N, d_{ij}) = -\alpha \cdot \log_2 N \cdot d_{ij} $, where $d_{ij}$ is pairwise distance and $\alpha$ is a learnable parameter. In our case, we first apply our initial aggregation layer to form a latent distance matrix of size $N \times N \times 32$. Then, for each ICAM layer, we introduce a learnable linear projection that compresses the vector of size 32 for each node pair into a scalar, which is used instead of $d_{ij}$. The linear projection weights are different for each ICAM layer to enable more informative processing through the encoder stack. Embedded node features, if they are present, are used as ``row'' features. We utilize the same decoder as in the original paper in the single-objective case, namely an AAFM-based architecture. 

\subsection{Training Pseudo-Code}
\label{app: pseudo-code}
Algorithm~\ref{alg: training} presents our hierarchical training setup. In the multi-objective case, we additionally sample one preference $\lambda \sim \Lambda$ per batch and use the scalarized reward $R_\lambda = -C_\lambda$. 

\begin{algorithm}[t]
\caption{One batch of NEPF hierarchical training}
\label{alg: training}
\begin{algorithmic}[1]
\Require instance distribution $\mathcal{G}$, batch size $B$, POMO size $K_1$, edge samples $K_2$, learning rate $\eta$
  \State sample instances $G_1, \dots, G_B \sim \mathcal{G}$
  \For{$i = 1$ \textbf{to} $B$} \Comment{parallelized in practice}
    \State $D_i \gets \textsc{Aggregate}(G_i)$ \Comment{pre-encoding, \eqref{eq: pre-encoding}}
    \State $\{h_u\} \gets \textsc{Encode}(D_i)$
    \For{$j = 1$ \textbf{to} $K_1$} \Comment{POMO permutations, parallelized in practice}
      \State $\boldsymbol{\pi}_{ij} \sim p^\text{node}_{\theta_1}(\cdot \mid G_i)$
      \For{$k = 1$ \textbf{to} $K_2$} \Comment{one-shot, sampled in parallel}
        \State $\boldsymbol{\epsilon}_{ijk} \sim p^\text{edge}_{\theta_2}(\cdot \mid \boldsymbol{\pi}_{ij}, G_i)$
        \State $R(\boldsymbol{\pi}_{ij}, \boldsymbol{\epsilon}_{ijk}) \gets -C(\boldsymbol{\pi}_{ij}, \boldsymbol{\epsilon}_{ijk})$
      \EndFor
      \State $b^\text{edge}(\boldsymbol{\pi}_{ij}, G_i) \gets \frac{1}{K_2}\sum_{k} R(\boldsymbol{\pi}_{ij}, \boldsymbol{\epsilon}_{ijk})$
      \State $R^*(\boldsymbol{\pi}_{ij}) \gets \max_{k} R(\boldsymbol{\pi}_{ij}, \boldsymbol{\epsilon}_{ijk})$
    \EndFor
    \State $b^\text{node}(G_i) \gets \frac{1}{K_1}\sum_{j} R^*(\boldsymbol{\pi}_{ij})$
  \EndFor
  \State estimate $\nabla_\theta J(\theta)$ \Comment{\eqref{eq: reinforce}; optionally add $\nabla_\theta \mathcal{L}_\text{est}$}
  \State $\theta \gets \textsc{Adam}(\theta, \nabla_\theta J(\theta), \eta)$
\end{algorithmic}
\end{algorithm}

\section{Instance Generation, Problems and Baselines}
\label{app: problems}

\subsection{Multigraph Distributions}
To evaluate under controlled conditions, we adopt the two synthetic multigraph distributions FLEX$x$ and FIX$x$ introduced by \citet{rydin2026beyond}. Both these distributions treat node pairs independently. 

\paragraph{FLEX$x$.} For each edge, this distribution samples the edge attributes independently $\sim \text{Unif}[0, 1]$. The generation starts by sampling $x$ such edges between each node pair. Then, dominated edges are pruned, leaving a Pareto set of edges connecting each node pair. Thus, this distribution has a flexible number of edges, with at most $x$ per node pair. 

\paragraph{FIX$x$.} This distribution starts by sampling $x$ edges per node pair independently in the same way as FLEX$x$. Then, instead of pruning dominated edges, the generation proceeds by sorting the attributes among the parallel edges. Specifically, considering two attributes (our setting), one attribute is sorted in increasing order and the other in decreasing order. Thus, the edge with lowest value in the first attribute will have the highest value in the second attribute etc. This leaves exactly $x$ edges between each node pair, with perfectly negatively correlated attributes. 

\subsection{RCTSP}

In the resource constrained TSP \citep{RCTSP}, each edge is associated with a cost and a resource consumption. The total cost and resource consumption are given by the sum over the traversed edges. The goal is to minimize the cost of a tour visiting all nodes, while ensuring the total resource consumption stays under a threshold $R$. 

\paragraph{Data generation.} For a fixed distribution, edge count and node count, the resource constraint $R$ is constant across instances. Let $R_\text{cost}$ and $R_\text{resource}$ be the expected resource consumption for a cost-optimal and resource-optimal tour respectively. We set 
\begin{equation}
\label{eq: resource}
    R = (R_\text{cost} + R_\text{resource}) / 4.
\end{equation}
This constraint tightness level ensures the problem does not reduce into a standard TSP, while it is generally not difficult to find feasible solutions. 

\paragraph{Model details.} Edge features are the cost and resource consumption and there are no node features. In the decoding for all models, we augment the context with the current resource usage $r_t$ in \eqref{eq: context} by setting $s_t = r_t$. To encourage feasibility, we subtract the constraint violation multiplied with a constant factor from the reward during training. This leads to no recorded infeasible instances for any model during evaluation in our experiments.

\paragraph{Baselines.} Besides Gurobi (with no time limit, applied to the problem formulation in Appendix~\ref{app: MILPs}), we design a beam search heuristic as baseline. It incrementally constructs tours using a Lagrangian score (cost + $\lambda \cdot$resource). At each step, partial tours are pruned via feasibility checks and Pareto dominance, retaining a fixed number of beam states. Completed tours are repaired if needed by selecting alternative parallel edges with lower resource consumption. The trade-off parameter $\lambda$ is adaptively updated across a fixed number of outer iterations to balance cost and resource feasibility.

\subsection{OP}
\label{app: OP}

Our second single-objective problem is a multi-constraint variant of the orienteering problem. Each edge is associated with two costs and the sum of both costs over traversed edges must not exceed fixed thresholds $T_1$ and $T_2$. The goal is to maximize the sum of collected prizes over the visited nodes. As in standard orienteering formulations, not all nodes need to be visited, and the route must terminate at the depot (see \citet{OP}).

\paragraph{Data generation.} Prizes are sampled uniformly in $[0, 1]$ independently for each node. Similarly to the RCTSP, we set thresholds using the average incurred costs of optimal paths. Let $C^{(j)}_i$ be expected cost $i$ for a route that is optimal with respect to cost $j$. Since our graph distributions are symmetric with respect to the costs, we note that $C^{(2)}_1 = C^{(1)}_2$ and $C^{(1)}_1 = C^{(2)}_2$. The thresholds are thus set to 
\begin{equation}
    \label{eq: thresholds}
    T_1 = T_2 = (C^{(1)}_2 + C^{(2)}_2) / 8 = (C^{(1)}_1 + C^{(2)}_1) / 8,
\end{equation}
where the denominator ensures that many nodes can be visited, but not all. The exception is the FLEX$5$ distribution, where the above threshold is too lenient, allowing all nodes to be visited. For this case, we set the threshold to half the value in \eqref{eq: thresholds}.

\paragraph{Model details.} 
Importantly, since neither graph distribution satisfies the triangle inequality, it is non-trivial to mask out infeasible nodes in the decoding. As such, following \citet{lischkagreat}, we let the models learn when to return to the depot. We do this by penalizing constraint violation during training as with the RCTSP.

Besides the two-dimensional edge costs, we concatenate the node prize to each incident edge to form edge features in the first stage. In the FSASP stage, we only utilize the original edge costs as features, as prizes are not needed. In the node permutation MP decoder, we augment the context with the current cost level $c_t \in \mathbb{R}^2$ by setting $s_t = c_t$.

\paragraph{Baselines.} The main classical baseline is Gurobi, applied without time limit to the problem formulation in Appendix~\ref{app: MILPs}. We also design a greedy constructive heuristic that iteratively builds a path starting from the depot. At each step, it selects the next node by maximizing prize-to-cost ratio, where edge costs are weighted by remaining resource until threshold is reached. A candidate move is only considered if it admits a direct feasible return path to the depot. The process terminates when no feasible extension exists.

\subsection{MOTSP}
As the simplest multi-objective problem, we consider the multi-objective TSP. Here, edge costs are two-dimensional, and the goal is to find the Pareto set of routes minimizing the two-dimensional sum over all edges. 

\paragraph{Data generation.} This problem has no additional features beyond the edge costs, which we sample as usual from FLEX$x$ and FIX$x$.

\paragraph{Model details.} For this problem, we do not utilize the FSASP stage. Instead, we replace it by always selecting the parallel edge with the lowest linearly scalarized cost. This reduces runtime while maintaining performance. 

\paragraph{Baselines.} We consider a nearest-neighbor heuristic as well as LKH and Google OR-tools. These utilize linear preference-based scalarization to transform the multi-objective problem into several single-objective problems. For each scalarized problem, the multigraph is pruned by removing sub-optimal edges with respect to the linearly scalarized cost. We sweep across 101 preferences to form the Pareto front for each problem instance. LKH uses default parameters (no explicit time limit) and OR-tools uses the ``PATH\_CHEAPEST\_ARC'' strategy and default greedy-descent local search, run to its first local optimum. 

\paragraph{Reference for HV calculation.} We use $(60, 60)$ and $(100, 100)$ as reference for the FLEX$x$-100 and FIX$x$-100 distributions respectively, regardless of $x$. In the fine-tuned experiments in Table~\ref{tab: fine_tuned_motsp_app} (and Table~\ref{tab: fine_tuned_motsp} in the main text), we use $(120, 120)$ for FLEX$2$-200, $(180, 180)$ for FLEX$10$-300, $(300, 300)$ for FLEX$2$-500 and $(600, 600)$ for FLEX$2$-1000 while we use $(N, N)$ for FIX$x$-$N$.

\subsection{MOCVRP}
Similarly to the MOTSP, the multi-objective capacitated vehicle routing problem features two edge costs, which are summed over the route to yield the objectives. In this case, however, the vehicle must respect a capacity $c$, while satisfying customer demands $d_i, \: i \in V$. It starts at a depot, and must return occasionally to drop off its load. The route finishes when all customers have been visited.

\paragraph{Data generation.} Similarly to \citet{Kool2018AttentionLT}, we set the vehicle capacity to $c=50$ and sample demands uniformly and independently from the set $\{1, \dots, 9\}$.

\paragraph{Model details.} Edge features are the two edge costs as well as demand of the end-point node. As in the MOTSP, we do not utilize the FSASP stage for this problem, instead always selecting the edge that minimizes linearly scalarized cost. Finally, we augment the context with the current vehicle load $l_t$ by setting $s_t = l_t$ in the MP decoder.

\paragraph{Baselines.} We utilize a nearest-neighbor heuristic, HGS and Google OR-tools as baselines. Similarly to the MOTSP, to make these baselines applicable we use linear scalarization across 101 preferences together with multigraph pre-pruning. HGS uses default parameters (no explicit time limit) and OR-tools uses the same setup as for the MOTSP. 

\paragraph{Reference for HV calculation.} We consider the same reference points as for the MOTSP.

\subsection{MOTSPTW}
Following \citet{rydin2026beyond}, we also benchmark on the multi-objective TSP with time-windows. In this problem, one objective is the number of violated time windows, while the other is the traveled distance. Each edge is associated with a distance as well as a travel time. The vehicle starts at a depot and must visit all customers. It must immediately leave a customer upon arrival. That is, it cannot wait until a time window starts.

As this problem features edge attributes interacting with node attributes and early edge selections impact downstream penalties, solving the FSASP is crucial for satisfactory performance.  

\paragraph{Data generation.} We sample time-windows according to the ``medium'' distribution of \citet{chen_looking_2024} and \citet{BiLearningToHandle}, following the setup in \citet{rydin2026beyond}.

\paragraph{Model details.} Edge features in both stages are the original edge features (distance, travel time) concatenated with the time-window of the end-point node.

\paragraph{Baseline.} We use an insertion heuristic for the linearly scalarized problem. Nodes are iteratively inserted at the position that minimizes the preference weighted sum of incremental distance and time-window violation. For each insertion, all parallel edge options are considered, and downstream arrival times are updated accordingly.

\paragraph{Reference for HV calculation.} For FLEX$x$, we utilize $(105, 60)$ as reference, where the first objective is the time-window violation. For FIX$x$, we utilize $(105, 100)$.

\subsection{MOOP}
Our last problem is a bi-objective version of the orienteering problem. In this setting, each edge is associated with a cost and a resource. One objective is the sum of costs of traversed edges (to be minimized), whereas the other objective is the sum of collected prizes across visited nodes (to be maximized). The resource consumption must not exceed a threshold $R$, which ensures the vehicle might not be able to visit all nodes. Note that there is an inherent trade-off between choosing low-cost edges (that minimize the cost objective) and low-resource edges (which ensure the vehicle can visit more nodes). 

As we prefer a pure minimization problem for compatibility with our other problems, we change the implementation slightly. Instead of maximizing collected prizes our implementation minimizes prizes not collected.  

\paragraph{Data generation.} We sample prizes as in the OP (see Section~\ref{app: OP}) and set the resource limit as in the RCTSP (see \eqref{eq: resource}).

\paragraph{Model details.} Edge features in the node permutation stage are cost, resource consumption and end-point node prize. In the FSASP stage, edge features are only cost and resource consumption. Additionally, we augment the context with current resource consumption $r_t$. As in the OP, the model needs to learn when to return to the depot. Thus we set the collected prize to zero when the resource constraint is violated. 

\paragraph{Baseline.} Similarly to the OP, we use a greedy heuristic which maximizes prize-to-cost ratio of the next edge traversal. We run this heuristic for each multi-objective preference vector $\lambda = (\lambda_1, \lambda_2)$. For a single run, the prize in the numerator is weighted with the preference of the prize objective $\lambda_1$ and the cost in the denominator is weighted by $\lambda_2$.

\paragraph{Reference for HV calculation.} For FLEX$x$, we utilize $(50, 25)$ as reference (first objective is prize not collected). For FIX$x$, we utilize $(50, 30)$.

\subsection{MILP Formulations.}
\label{app: MILPs}

This section formalizes our problems by presenting their Mixed-Integer Linear Program (MILP) formulations. We refrain from formulating the MOTSP and MOCVRP as they are rather standard extensions of the TSP and CVRP to the multi-objective and multigraph settings. 

\paragraph{Common notation and constraints.} All problems are defined on the directed multigraph $G=(V,E)$ of Section~\ref{sec: notation}, with parallel-edge sets $E_{uv}$ between $(u,v)\in V^2$ and per-edge attributes $e_l$. We designate a distinguished node $0\in V$ (the depot or, for the RCTSP, an arbitrary reference node) and write $V^-:=V\setminus\{0\}$. The shared decision variables are the binary edge indicators $x_l\in\{0,1\}$, $l\in E$ ($x_l=1$ iff edge $l$ is traversed). 

Every model uses the degree constraints
\begin{align}
\sum_{u\in V\setminus\{v\}}\ \sum_{l\in E_{uv}} x_l = b_v, \qquad \sum_{v\in V\setminus\{u\}}\ \sum_{l\in E_{uv}} x_l = b_u, \qquad \forall\, u,v\in V, \label{eq:common-deg}
\end{align}
where the right-hand side is $b_u=1$ when all nodes must be visited (RCTSP, MOTSPTW) and $b_u=z_u$ with an added visit indicator $z_u\in\{0,1\}$ and $z_0=1$ when the visited set is itself a decision (OP, MOOP). Note that \eqref{eq:common-deg} also guarantees that at most one edge per node pair is selected. 

Unless a problem provides its own ordering (see the MOTSPTW below), subtours are removed by Miller--Tucker--Zemlin constraints on $V^-$, with ordering variables $\mu_u$,
\begin{align}
\mu_u-\mu_v+1 \ \le\ (|V|-1)\Big(1-\textstyle\sum_{l\in E_{uv}} x_l\Big), \quad \forall\, u,v\in V^-,\ u\neq v, \qquad 1\le\mu_u\le |V|-1. \label{eq:common-mtz}
\end{align}

\paragraph{RCTSP.}
Each edge has attributes $e_l=(c_l,\rho_l)$, a cost $c_l\ge 0$ and a resource consumption $\rho_l\ge 0$. The tour visits all nodes and its total resource may not exceed the budget $R$ of \eqref{eq: resource}. This is formulated as
\begin{align}
\min_{x,\mu}\ & \sum_{l\in E} c_l\,x_l \label{eq:rctsp} \\
\text{s.t.}\ & \eqref{eq:common-deg}\ \text{with}\ b_u=1, \\
& \sum_{l\in E} \rho_l\,x_l \le R, \\
& \eqref{eq:common-mtz}. \nonumber
\end{align}

\paragraph{OP.}
Each edge carries two costs $e_l=(c_l^{(1)},c_l^{(2)})$ and each customer $u\in V^-$ a prize $n_u=p_u\ge 0$. The route visits a chosen subset of nodes, terminates at the depot and must respect both cost thresholds $T_1, T_2$ of \eqref{eq: thresholds}, while maximizing collected prize:
\begin{align}
\max_{x,z,\mu}\ & \sum_{u\in V^-} p_u\,z_u \label{eq:op} \\
\text{s.t.}\ & \eqref{eq:common-deg}\ \text{with}\ b_u=z_u,\ z_0=1, \\
& \sum_{l\in E} c_l^{(1)} x_l \le T_1, \\
& \sum_{l\in E} c_l^{(2)} x_l \le T_2, \\
& \eqref{eq:common-mtz}. \nonumber
\end{align}

\paragraph{MOTSPTW.}
Each edge carries $e_l=(\delta_l,\tau_l)$, a distance $\delta_l\ge 0$ and a travel time $\tau_l\ge 0$, and each customer a time window $n_u=[a_u,b_u]$. Apart from $x$, we use arrival times $y_u\ge 0$ and violation indicators $w_u\in\{0,1\}$, $u\in V^-$ as decision variables. The two costs are the number of violated windows and the traveled distance:
\begin{align}
\min_{x,y,w}\ & \Big(\ \textstyle\sum_{u\in V^-} w_u,\ \ \sum_{l\in E}\delta_l\,x_l\ \Big) \label{eq:motsptw-obj} \\
\text{s.t.}\ & \eqref{eq:common-deg}\ \text{with}\ b_u=1,\qquad y_0=0, \label{eq:motsptw-deg} \\
& y_v \ \ge\ y_u + \tau_l - M(1-x_l), \label{eq:motsptw-prop1} \\
& y_v \ \le\ y_u + \tau_l + M(1-x_l), && \forall\, (u,v)\in V^2,\ u\neq v,\ v\neq 0,\ l\in E_{uv}, \label{eq:motsptw-prop2} \\
& a_u - M\,w_u\ \le\ y_u\ \le\ b_u + M\,w_u, && \forall\, u\in V^-. \label{eq:motsptw-viol}
\end{align}
On a used edge, the two inequalities~\eqref{eq:motsptw-prop1}--\eqref{eq:motsptw-prop2} become the equality $y_v=y_u+\tau_l$, which corresponds to the no-wait rule (the vehicle leaves immediately). Together with $y_0=0$ these constraints also act as generalized MTZ constraints and eliminate subtours whenever $\tau_l>0$. Constraint~\eqref{eq:motsptw-viol} forces $w_u=1$ whenever $y_u\notin[a_u,b_u]$, penalizing early or late arrival without affecting feasibility. 

\paragraph{MOOP.}
Each edge carries $e_l=(c_l,\rho_l)$, a cost $c_l\ge 0$ and a resource consumption $\rho_l\ge 0$, and each customer a prize $n_u=p_u\ge 0$. Visit indicators $z_u$ are as in the OP. As described above we minimize the prize not collected (equivalent to maximizing collected prize, since $\sum_{u}p_u$ is constant). The two costs are the uncollected prize and the traveled cost, subject to the resource budget $R$ of \eqref{eq: resource}:
\begin{align}
\min_{x,z,\mu}\ & \Big(\ \textstyle\sum_{u\in V^-} p_u(1-z_u),\ \ \sum_{l\in E} c_l\,x_l\ \Big) \label{eq:moop} \\
\text{s.t.}\ & \eqref{eq:common-deg}\ \text{with}\ b_u=z_u,\ z_0=1, \\
& \sum_{l\in E} \rho_l\,x_l \le R, \\
& \eqref{eq:common-mtz}. \nonumber
\end{align}

\section{Additional Experiments}

\subsection{Additional Single-Objective Results}
\label{app: SO}
Table~\ref{tab: so_app} shows results for our two single-objective problems on FLEX$5$ and FIX$5$. We note that NEPF is the best neural method on both problems with FLEX$5$, while it degrades on FIX$5$, especially compared to GMS-DH. Since FLEX$5$ and FIX$5$ share the same maximal number of parallel edges and only differ in the correlation structure, this indicates that the degradation stems from the adversarial attribute correlation of FIX$x$ (the attributes are perfectly negatively correlated, with extreme homogeneity among node pairs) rather than the edge count. In particular, on the more realistic multigraphs of Appendix~\ref{app: rw} (up to $22$ edges per node pair) and the real-world road networks of Appendix~\ref{app: real-world-instances}, NEPF remains the best neural method, suggesting that the FIX$5$ results on the single-objective RCTSP and OP reflect a synthetic stress test rather than a limitation of our factorization in practice. We also note that NEPF outperforms GMS on FIX$5$ for the MOTSP, MOCVRP and MOTSPTW (see Table~\ref{tab: MO}).

\begin{table*}
    \centering
    \caption{Single-objective results on the FLEX5-100 and FIX5-100 distributions (five parallel edges per node pair).}
    \label{tab: so_app}
    {
    \setlength{\tabcolsep}{0.07cm}
    \renewcommand{\arraystretch}{0.8}
    \fontsize{8}{11}\selectfont
    \begin{tabular}{lcccccccccccc}
\toprule
\multicolumn{1}{l}{}
& \multicolumn{6}{c|}{\textbf{RCTSP}}
& \multicolumn{6}{c}{\textbf{OP}} \\
\multicolumn{1}{l}{}
& \multicolumn{3}{c}{FLEX5-100} & \multicolumn{3}{c|}{FIX5-100}
& \multicolumn{3}{c}{FLEX5-100} & \multicolumn{3}{c}{FIX5-100} \\
\multicolumn{1}{l}{}
& Obj. & Gap & \multicolumn{1}{c}{Time}
& Obj. & Gap & \multicolumn{1}{c|}{Time}
& Obj. & Gap & \multicolumn{1}{c}{Time}
& Obj. & Gap & Time \\
\midrule
\multicolumn{1}{l|}{Gurobi}
& 1.49 & 0.00\% & \multicolumn{1}{c|}{(9.2h)}
& 14.4 & 0.00\% & \multicolumn{1}{c|}{(7.8h)}
& 41.1 & 0.00\% & \multicolumn{1}{c|}{(5.3h)}
& 39.3 & 0.00\% & (16h) \\
\multicolumn{1}{l|}{Beam/Greedy}
& 3.74 & 151.50\% & \multicolumn{1}{c|}{(2.7h)}
& 18.8 & 30.55\% & \multicolumn{1}{c|}{(3.0h)}
& 30.6 & 25.47\% & \multicolumn{1}{c|}{(1.2m)}
& 34.2 & 12.80\% & (1.2m) \\
\cmidrule(lr){2-13}
\multicolumn{1}{l|}{GMS-EB}
& 2.24 & 50.60\% & \multicolumn{1}{c|}{(5.4m)}
& 27.4 & 90.14\% & \multicolumn{1}{c|}{(7.2m)}
& 37.5 & 8.71\% & \multicolumn{1}{c|}{(4.7m)}
& 37.2 & 5.16\% & (6.3m) \\
\multicolumn{1}{l|}{GMS-EB (aug)}
& \underline{2.07} & \underline{39.01\%} & \multicolumn{1}{c|}{(42m)}
& 26.4 & 83.07\% & \multicolumn{1}{c|}{(59m)}
& 37.5 & 8.70\% & \multicolumn{1}{c|}{(38m)}
& \underline{37.2} & \underline{5.16\%} & (48m) \\
\multicolumn{1}{l|}{GMS-DH}
& 2.43 & 63.15\% & \multicolumn{1}{c|}{(1.8m)}
& 16.1 & 11.65\% & \multicolumn{1}{c|}{(3.9m)}
& 38.6 & 6.05\% & \multicolumn{1}{c|}{(1.7m)}
& 37.8 & 3.78\% & (4.0m) \\
\multicolumn{1}{l|}{GMS-DH (aug)}
& 2.19 & 47.23\% & \multicolumn{1}{c|}{(14m)}
& \textbf{15.8$^\star$} & \textbf{9.52\%} & \multicolumn{1}{c|}{(31m)}
& \underline{38.6} & \underline{6.05\%} & \multicolumn{1}{c|}{(14m)}
& \textbf{37.8$^\star$} & \textbf{3.78\%} & (29m) \\
\cmidrule(lr){2-13}
\multicolumn{1}{l|}{NEPF}
& 1.84 & 23.41\% & \multicolumn{1}{c|}{(43s)}
& 21.3 & 48.27\% & \multicolumn{1}{c|}{(53s)}
& 38.7 & 5.88\% & \multicolumn{1}{c|}{(43s)}
& 34.3 & 12.55\% & (48s) \\
\multicolumn{1}{l|}{NEPF (aug)}
& \textbf{1.74$^\star$} & \textbf{16.58\%} & \multicolumn{1}{c|}{(5.9m)}
& \underline{20.6} & \underline{43.10\%} & \multicolumn{1}{c|}{(7.1m)}
& \textbf{38.7$^\star$} & \textbf{5.87\%} & \multicolumn{1}{c|}{(5.9m)}
& 34.3 & 12.54\% & (6.7m) \\
\bottomrule
\end{tabular}
    }
\end{table*}

\subsection{Fine-Tuned Results on Larger Instances}
\label{app: fine_tuned}
\paragraph{Experimental details.} First, we remark that the poor memory and runtime scaling of the baselines means that fully trained experiments on larger graphs are infeasible in general. Hence, we only consider fine-tuned experiments, where the models trained for $N=100$ are tuned for a short time on larger graph distributions. When even such training is infeasible due to memory constraints, we instead consider zero-shot generalization, without any fine-tuning. 

For GMS-DH and NEPF, we tune each model for 10 epochs with 10\,000 instances per epoch randomly drawn from the target distribution. With GMS-EB we use only 2000 instances per epoch over 10 epochs, owing to its considerable training time on large graphs. We report fine-tuned HV and corresponding optimality gap, as well as zero-shot HV and gap and the total time for the fine-tuning. Tables~\ref{tab: fine_tuned_motsp} and \ref{tab: fine_tuned_mocvrp} in the main text report fine-tuned HV and gap, unless those metrics are unavailable, in which case zero-shot HV and gap are reported. The models we fine-tune are those trained for settings closest to the target distribution. For example, for FIX$10$-300 we tune FIX$5$-100, and for FLEX2-500 we tune FLEX2-100. 

\paragraph{Further results.} Tables~\ref{tab: fine_tuned_motsp_app} and \ref{tab: fine_tuned_mocvrp_app} display further generalization results for the MOTSP and MOCVRP respectively. For the MOTSP, GMS-EB consistently outperforms NEPF zero-shot. However, fine-tuning NEPF, which is generally infeasible for the baselines, makes it the best of the neural methods. For the MOCVRP, GMS-EB generally degrades significantly in the zero-shot setting, and GMS-DH is the strongest baseline. Nevertheless, NEPF obtains the best performance in both the zero-shot and fine-tuned settings in 5/6 cases for this problem. 

Last, we note that scaling neural methods to larger instance sizes for multigraphs is inherently more difficult than the Euclidean case. Multigraph routing is an edge-based task, requiring edge-level representation and actions, while Euclidean routing often works on node-level representations. The instance sizes we consider should thus not be compared to the state-of-the-art sizes for Euclidean routing, which are considerably higher \citep{ye2024glop}.

\begin{table*}
    \centering
    \caption{Fine-tuned results on larger instances for the MOTSP. For all distributions and instance sizes, we report Fine-Tuned (FT) and Zero-Shot (ZS) HV together with corresponding gaps. We also report the total time for the fine-tuning, marking with OOM when such tuning is infeasible due to memory constraints.}
    \label{tab: fine_tuned_motsp_app}
    {
    \setlength{\tabcolsep}{0.07cm}
    \renewcommand{\arraystretch}{0.8}
    \fontsize{8}{11}\selectfont
    \begin{tabular}{lcccccccccc}
\toprule
\multicolumn{1}{l|}{}
& \multicolumn{5}{c|}{FLEX2-200}
& \multicolumn{5}{c}{FIX2-200} \\
\multicolumn{1}{l|}{}
& FT HV & FT Gap & ZS HV & ZS Gap & \multicolumn{1}{c|}{FT Time} & FT HV & FT Gap & ZS HV & ZS Gap & FT Time \\
\midrule
\multicolumn{1}{l|}{LKH}   & - & - & 0.97 & 0.00\% & \multicolumn{1}{c|}{-} & - & - & 0.97 & 0.00\% & - \\
\multicolumn{1}{l|}{GMS-EB}& 0.95 & 1.93\% & 0.95 & 1.63\% & \multicolumn{1}{c|}{7.4h} & 0.95 & 1.89\% & 0.96 & 1.20\% & 5.9h \\
\multicolumn{1}{l|}{GMS-DH}& 0.95 & 1.97\% & 0.95 & 2.37\% & \multicolumn{1}{c|}{5.8h} & 0.94 & 2.97\% & 0.94 & 2.68\% & 6.8h \\
\multicolumn{1}{l|}{NEPF}  & 0.95 & 1.42\% & 0.95 & 1.70\% & \multicolumn{1}{c|}{2.8h} & 0.96 & 1.07\% & 0.96 & 1.24\% & 2.9h \\
\midrule
\multicolumn{1}{l|}{}
& \multicolumn{5}{c|}{FLEX10-300}
& \multicolumn{5}{c}{FIX10-300} \\
\midrule
\multicolumn{1}{l|}{LKH}   & - & - & 0.99 & 0.00\% & \multicolumn{1}{c|}{-} & - & - & 0.89 & 0.00\% & - \\
\multicolumn{1}{l|}{GMS-EB}& - & - & 0.99 & 0.65\% & \multicolumn{1}{c|}{OOM} & - & - & 0.88 & 0.89\% & OOM \\
\multicolumn{1}{l|}{GMS-DH}& - & - & 0.97 & 1.99\% & \multicolumn{1}{c|}{OOM} & - & - & 0.87 & 2.48\% & OOM\\
\multicolumn{1}{l|}{NEPF}  & 0.99 & 0.38\% & 0.98 & 0.88\% & \multicolumn{1}{c|}{7.6h} & 0.88 & 0.84\% & 0.88 & 1.17\% & 9.8h\\
\midrule
\multicolumn{1}{l|}{}
& \multicolumn{5}{c|}{FLEX2-500}
& \multicolumn{5}{c}{FIX2-500} \\
\midrule
\multicolumn{1}{l|}{LKH}   & - & - & 0.98 & 0.00\% & \multicolumn{1}{c|}{-} & - & - & 0.98 & 0.00\% & -\\
\multicolumn{1}{l|}{GMS-EB}& - & - & 0.97 & 1.80\% & \multicolumn{1}{c|}{OOM} & - & - & 0.97 & 1.23\% & OOM\\
\multicolumn{1}{l|}{GMS-DH}& - & - & 0.95 & 3.20\% & \multicolumn{1}{c|}{OOM} & - & - & 0.95 & 2.92\% & OOM\\
\multicolumn{1}{l|}{NEPF}  & 0.97 & 1.33\% & 0.97 & 1.91\% & \multicolumn{1}{c|}{5.1h} & 0.97 & 1.11\% & 0.97 & 1.34\% & 5.0h\\
\midrule
\multicolumn{1}{l|}{}
& \multicolumn{5}{c|}{FLEX2-1000}
& \multicolumn{5}{c}{FIX2-1000} \\
\midrule
\multicolumn{1}{l|}{LKH}   & - & - & 0.99 & 0.00\% & \multicolumn{1}{c|}{-} & - & - & 0.99 & 0.00\% &  -\\
\multicolumn{1}{l|}{GMS-EB}& - & - & 0.97 & 1.62\% & \multicolumn{1}{c|}{OOM} & - & - & 0.98 & 1.14\% &  OOM\\
\multicolumn{1}{l|}{GMS-DH}& - & - & 0.95 & 4.08\% & \multicolumn{1}{c|}{OOM} & - & - & 0.96 & 2.94\% &  OOM\\
\multicolumn{1}{l|}{NEPF}  & - & - & 0.97 & 2.49\% & \multicolumn{1}{c|}{OOM} & - & - & 0.97 & 1.59\% &  OOM\\
\bottomrule
\end{tabular}
    }
\end{table*}

\begin{table*}
    \centering
    \caption{Fine-tuned results on larger instances for the MOCVRP. We do not use HGS as a baseline here, as its runtime becomes excessively high for these instance sizes.}
    \label{tab: fine_tuned_mocvrp_app}
    {
    \setlength{\tabcolsep}{0.07cm}
    \renewcommand{\arraystretch}{0.8}
    \fontsize{8}{11}\selectfont
    \begin{tabular}{lcccccccccc}
\toprule
\multicolumn{1}{l|}{}
& \multicolumn{5}{c|}{FLEX2-200}
& \multicolumn{5}{c}{FIX2-200} \\
\multicolumn{1}{l|}{}
& FT HV & FT Gap & ZS HV & ZS Gap & \multicolumn{1}{c|}{FT Time} & FT HV & FT Gap & ZS HV & ZS Gap & FT Time \\
\midrule
\multicolumn{1}{l|}{GMS-EB}& 0.77 & 14.59\% & 0.51 & 42.85\% & \multicolumn{1}{c|}{14h} & 0.88 & 1.66\% & 0.82 & 7.59\% & 13h\\
\multicolumn{1}{l|}{GMS-DH}& 0.87 & 3.88\% & 0.86 & 4.08\% & \multicolumn{1}{c|}{6.3h} & 0.89 & 0.00\% & 0.89 & 0.39\% & 10h\\
\multicolumn{1}{l|}{NEPF}  & 0.90 & 0.00\% & 0.90 & 0.60\% & \multicolumn{1}{c|}{4.4h} & 0.88 & 1.41\% & 0.87 & 2.59\% & 6.0h\\
\midrule
\multicolumn{1}{l|}{}
& \multicolumn{5}{c|}{FLEX10-300}
& \multicolumn{5}{c}{FIX10-300} \\
\midrule
\multicolumn{1}{l|}{GMS-EB}& - & - & 0.05 & 94.38\% & \multicolumn{1}{c|}{OOM} & - & - & 0.00 & 100.00\% & OOM\\
\multicolumn{1}{l|}{GMS-DH}& - & - & 0.84 & 13.55\% & \multicolumn{1}{c|}{OOM} & - & - & 0.69 & 13.63\% & OOM\\
\multicolumn{1}{l|}{NEPF}  & 0.97 & 0.00\% & 0.97 & -0.03\% & \multicolumn{1}{c|}{18h} & 0.80 & 0.00\% & 0.79 & 1.32\% & 19h\\
\midrule
\multicolumn{1}{l|}{}
& \multicolumn{5}{c|}{FLEX2-500}
& \multicolumn{5}{c}{FIX2-500} \\
\midrule
\multicolumn{1}{l|}{GMS-EB}& - & - & 0.01 & 99.34\% & \multicolumn{1}{c|}{OOM} & - & - & 0.14 & 84.05\% & OOM\\
\multicolumn{1}{l|}{GMS-DH}& - & - & 0.73 & 21.56\% & \multicolumn{1}{c|}{OOM} & - & - & 0.86 & 3.61\% & OOM\\
\multicolumn{1}{l|}{NEPF}  & 0.93 & 0.00\% & 0.92 & 1.02\% & \multicolumn{1}{c|}{35h} & 0.89 & 0.00\% & 0.89 & 0.13\% & 36h\\
\bottomrule
\end{tabular}
    }
\end{table*}

\subsection{Scalability Stress Test}
\label{app: stress_test}

Table~\ref{tab:stress_test} stress tests the scalability of NEPF, compared to the neural baselines. With the current hardware, it is possible to solve instances up to $N=2500$ with $M=2$ and $N=2000$ with $M=5$. In contrast, both GMS variants run out of memory beyond $N=1000$ at either edge count, corresponding to a circa $2.5\times$ extension of the maximum feasible node count and a circa $5\times$ extension of the maximum total edge count. The remaining bottleneck is the one-time pre-encoding aggregation, which, being independent across node pairs, admits chunked processing. We leave this optimization for future work. The same analysis is reflected in runtime. NEPF's per-instance time is essentially invariant to the number of parallel edges ($50.0$s vs.\ $47.3$s at $N=1000$ for $M=5$ vs.\ $M=2$), while GMS-DH slows down by a factor ${\approx}2.4$ and GMS-EB by a factor ${\approx}4$ when moving from $M=2$ to $M=5$. Thus, the advantage of NEPF grows with the density of the multigraph.

\begin{table}[t]
\centering
\caption{Scalability stress test on the MOTSP. Each cell reports per-instance
inference time / peak GPU memory. Protocol: batch size $B=1$, $K_1=100$ POMO
samples, half precision, single NVIDIA Tesla A40 (48\,GB). OOM denotes exceeding
48\,GB. $^{(a)}$GMS-EB failed at this size due to implementation issues, not memory constraints.}
\label{tab:stress_test}
\setlength{\tabcolsep}{3.5pt}
\small
\begin{tabular}{lcccccc}
\toprule
& \multicolumn{6}{c}{Number of nodes $N$} \\
\cmidrule(lr){2-7}
Method & 100 & 200 & 500 & 1000 & 2000 & 2500 \\
\midrule
\multicolumn{7}{l}{$M = 2$} \\
\midrule
NEPF   & 5.3s / 0.08GB  & 9.9s / 0.24GB  & 25s / 1.40GB & 47s / 5.49GB   & 97s / 21.8GB & 120s / 33.2GB \\
GMS-DH & 5.2s / 0.13GB & 9.8s / 0.45GB  & 25s / 2.66GB & 57s / 10.6GB  & OOM & OOM \\
GMS-EB & 5.8s / 0.12GB & 12s / 0.43GB & 35s / 2.53GB & 130s / 10.0GB & OOM & OOM \\
\midrule
\multicolumn{7}{l}{$M = 5$} \\
\midrule
NEPF   & 5.2s / 0.14GB  & 9.7s / 0.49GB   & 24s / 3.01GB  & 50s / 11.9GB  & 91s / 43.1GB & OOM \\
GMS-DH & 11s / 0.28GB & 21s / 1.07GB & 57s / 6.61GB  & 140s / 26.4GB & OOM & OOM \\
GMS-EB & 12s / 0.27GB & 27s / 1.03GB & 140s / 6.29GB & ---$^{(a)}$ & OOM & OOM \\
\bottomrule
\end{tabular}
\end{table}

\subsection{Runtime-Quality Trade-Off}
\label{app: runtime_quality}
This section features further analysis regarding the runtime-quality trade-off between NEPF and strong classical baselines with tuneable solving times. Specifically, we show that NEPF obtains higher quality solutions when the classical algorithms are constrained to similar computational budgets, demonstrating the usefulness of our approach. Also note that warm-starting the classical algorithms with NEPF-generated solutions enables further speed-ups and potential optimality guarantees.

Figure~\ref{fig: runtime_quality} shows the results. As baselines, we use Gurobi for the RCTSP and OP, LKH for the MOTSP, and HGS for the MOCVRP. All reported times are per instance and NEPF is evaluated with batch size $B=1$, i.e., solving instances one by one. For Gurobi and HGS, we vary the wall-clock time limit per instance. For LKH, we instead control the budget via the RUNS and MAX\_TRIALS parameters, since hard time limits below the candidate-set initialization cause LKH to terminate without a solution, and additionally vary the number of preferences $P$ in the scalarization sweep.

For the RCTSP and OP, no competitive Gurobi operating point exists within two orders of magnitude of NEPF's runtime. At NEPF's operating point (${\approx}0.1$s per instance), Gurobi returns almost no feasible RCTSP solution and near-empty OP routes. Even with a budget of $10$s Gurobi remains infeasible on $13$--$29\%$ of RCTSP instances and $18$--$25\%$ below NEPF's OP objective, while NEPF maintains $100\%$ feasibility throughout.

For the MOCVRP, NEPF lies above the entire measured HGS curve. NEPF (aug) at ${\approx}9$--$10$s exceeds the quality of HGS at $120$s on both distributions.

The MOTSP is the only problem where a classical method offers a competitive trade-off. Notably, LKH's hypervolume is determined almost entirely by the number of preferences $P$. With the full $P=101$, even the minimal configuration (RUNS${}=1$, MAX\_TRIALS${}=1$) reaches its saturation hypervolume, while increasing the per-solve effort sixfold adds at most $0.0002$ HV. Below ${\approx}7$s (the regime shown in the figure), LKH must reduce $P$ and its hypervolume collapses to $0.75$--$0.88$, whereas NEPF delivers $0.93$ or more in this regime.

\begin{figure}
    \centering
    \definecolor{classicblue}{RGB}{72,120,168}
\definecolor{nepforange}{RGB}{232,133,61}
\definecolor{nepfred}{RGB}{192,57,43}

\pgfdeclareplotmark{nepfstar}{%
  \node[star, star point ratio=2.25, minimum size=7pt, inner sep=0pt,
        draw=black, line width=0.3pt, fill=nepforange] {};}
\pgfdeclareplotmark{augstar}{%
  \node[star, star point ratio=2.25, minimum size=7pt, inner sep=0pt,
        draw=black, line width=0.3pt, fill=nepfred] {};}

\begin{tikzpicture}[font=\footnotesize]
\begin{groupplot}[
  group style={
    group size=2 by 4,
    horizontal sep=1.3cm,
    vertical sep=1.05cm,
  },
  width=6.2cm, height=3.6cm,
  xmode=log,
  log ticks with fixed point,
  grid=both,
  grid style={line width=0.2pt, draw=gray!25},
  tick label style={font=\footnotesize},
  label style={font=\footnotesize},
  title style={font=\footnotesize},
  every axis plot/.append style={line width=0.9pt},
]

\nextgroupplot[title={FLEX2-100},
  ylabel={\shortstack{RCTSP (Gurobi)\\Objective ($\downarrow$)}},
  ymin=3.2, ymax=5.2,
  legend to name=tradeofflegend,
  legend columns=3,
  legend style={draw=none, font=\footnotesize, /tikz/every even column/.append style={column sep=0.5cm}},
]
\addplot[classicblue, mark=*, mark size=1.6pt] coordinates
  {(0.2,3.7166) (1,4.5626) (10,4.6024)};
\addlegendentry{Classical solver}
\addplot[only marks, mark=nepfstar] coordinates {(0.1126,4.2268)};
\addlegendentry{NEPF}
\addplot[only marks, mark=augstar]  coordinates {(0.1553,4.0571)};
\addlegendentry{NEPF (aug)}
\node[font=\tiny, classicblue, below right, inner sep=1pt] at (axis cs:0.2,3.7166) {4\%};
\node[font=\tiny, classicblue, above, inner sep=2pt] at (axis cs:1,4.5626) {33\%};
\node[font=\tiny, classicblue, above left, inner sep=1pt] at (axis cs:10,4.6024) {71\%};
\node[font=\tiny, nepforange, above right, inner sep=2pt] at (axis cs:0.1126,4.2268) {100\%};

\nextgroupplot[title={FIX2-100}, ymin=5.5, ymax=17]
\addplot[classicblue, mark=*, mark size=1.6pt] coordinates
  {(0.1,6.9894) (0.2,16.3) (1,13.4892) (10,10.7412)};
\addplot[only marks, mark=nepfstar] coordinates {(0.1081,11.5205)};
\addplot[only marks, mark=augstar]  coordinates {(0.1574,11.0035)};
\node[font=\tiny, classicblue, below right, inner sep=1pt] at (axis cs:0.1,6.9894) {1.5\%};
\node[font=\tiny, classicblue, right, inner sep=2pt] at (axis cs:0.2,16.3) {58.1 (3\%)};
\node[font=\tiny, classicblue, above right, inner sep=1pt] at (axis cs:1,13.4892) {43.5\%};
\node[font=\tiny, classicblue, above, inner sep=2pt] at (axis cs:10,10.7412) {87\%};
\node[font=\tiny, nepforange, above right, inner sep=2pt] at (axis cs:0.1081,11.5205) {100\%};

\nextgroupplot[ylabel={\shortstack{OP (Gurobi)\\Objective ($\uparrow$)}}]
\addplot[classicblue, mark=*, mark size=1.6pt] coordinates
  {(0.05,0.9859) (0.1,0.9859) (0.2,9.3175) (1,28.1519) (10,35.6596)};
\addplot[only marks, mark=nepfstar] coordinates {(0.0994,44.6791)};
\addplot[only marks, mark=augstar]  coordinates {(0.1660,44.6994)};

\nextgroupplot[]
\addplot[classicblue, mark=*, mark size=1.6pt] coordinates
  {(0.05,0.9859) (0.1,0.9859) (0.2,13.0168) (1,23.1770) (10,29.2682)};
\addplot[only marks, mark=nepfstar] coordinates {(0.0689,35.3478)};
\addplot[only marks, mark=augstar]  coordinates {(0.1446,35.3690)};

\nextgroupplot[ylabel={\shortstack{MOTSP (LKH)\\HV ($\uparrow$)}}, xtick={2,5,10}]
\addplot[classicblue, mark=*, mark size=1.6pt] coordinates
  {(2.17,0.7942) (4.58,0.8813) (6.60,0.9442)};
\addplot[only marks, mark=nepfstar] coordinates {(5.3335,0.9292)};
\addplot[only marks, mark=augstar]  coordinates {(5.5519,0.9338)};

\nextgroupplot[xtick={2,5,10}]
\addplot[classicblue, mark=*, mark size=1.6pt] coordinates
  {(2.15,0.7481) (4.24,0.8706) (7.60,0.9502)};
\addplot[only marks, mark=nepfstar] coordinates {(5.3038,0.9394)};
\addplot[only marks, mark=augstar]  coordinates {(5.4952,0.9422)};

\nextgroupplot[ylabel={\shortstack{MOCVRP (HGS)\\HV ($\uparrow$)}},
  xlabel={Per-instance time (s)}]
\addplot[classicblue, mark=*, mark size=1.6pt] coordinates
  {(5,0.8405) (10,0.8429) (20,0.8475) (60,0.8549) (120,0.8591)};
\addplot[only marks, mark=nepfstar] coordinates {(8.9893,0.8634)};
\addplot[only marks, mark=augstar]  coordinates {(9.2927,0.8693)};

\nextgroupplot[xlabel={Per-instance time (s)}, ymax=0.905]
\addplot[classicblue, mark=*, mark size=1.6pt] coordinates
  {(5,0.8800) (10,0.8816) (20,0.8845) (60,0.8897) (120,0.8925)};
\addplot[only marks, mark=nepfstar] coordinates {(9.4245,0.8927)};
\addplot[only marks, mark=augstar]  coordinates {(9.8104,0.8962)};

\end{groupplot}

\node[anchor=south] at ($(group c1r1.north east)!0.5!(group c2r1.north west) + (0,0.55cm)$)
  {\pgfplotslegendfromname{tradeofflegend}};
\end{tikzpicture}
    \caption{Comparison between NEPF and classical solvers at various runtime-quality trade-offs. For the RCTSP, the number beside each data point indicates feasibility rate, which should be interpreted as the main metric rather than the objective value.}
    \label{fig: runtime_quality}
\end{figure}

\subsection{Realistic Instances}
\label{app: rw}
In this section, we show results when applying NEPF to more realistic instances zero-shot. The purpose is both to show that our method performs well out-of-distribution without any fine-tuning and to show that performance is maintained on distributions where the triangle inequality is satisfied. 

We generate instances by first sampling points uniformly in the unit square $[0, 1]^2$ and calculating Euclidean distances. Then, following \citet{LETCHFORD2014} and \citet{ben_ticha_empirical_2017}, we transform these Euclidean instances into multigraphs with the following procedure. First, for each node pair $(u, v)$ with Euclidean distance $d^{(1)}_{uv}$, we compute a second distance $d^{(2)}_{uv} = \nu \cdot d^{(1)}_{uv} + \mu \cdot \gamma_{uv}\cdot \hat{d}^{(1)}$, where: 
\begin{itemize}
    \item $\hat{d}^{(1)} = \max_{(u, v) \in V^2} d^{(1)}_{uv}$ is the maximum Euclidean distance.
    \item $\gamma_{uv} \sim \text{Unif}[0,1]$ is a random variable.
    \item $\mu$ and $\nu$ are parameters controlling the correlation between the distances $d^{(1)}$ and $d^{(2)}$. 
\end{itemize}
For the latter, we define three cases: Strong Correlation (SC): $(\nu, \mu) = (0.9, 0.1)$, Weak Correlation (WC): $(\nu, \mu) = (0.5, 0.5)$ and No Correlation (NC): $(\nu, \mu) = (0.1, 0.9)$. Finally, in the resulting simple graph with two edge distances, we solve a multi-objective shortest path problem for each node pair using dynamic programming. The output is a multigraph, where both edge distances satisfy the triangle inequality and where the number of edges between node pairs varies. Table~\ref{tab: real_stat} summarizes the statistics of instances generated with this method for 100 nodes. It shows the maximum number of edges per node pair, mean edges per node pair and mean edges per graph. In particular, remark that there exist node pairs with relatively large edge counts.  

Table~\ref{tab: real} shows results on these instances for one single-objective problem (RCTSP) and one multi-objective problem (MOTSP). We use the same baselines as in the main results in Section~\ref{sec: results}. All neural methods are applied zero-shot without training on similar instances (we apply the versions trained on FLEX5-100). Also note that we set a time limit for Gurobi to 6~minutes per instance for these experiments.

To summarize these results, NEPF consistently shows strong zero-shot generalization. Among the neural methods, it performs best in 5/6 cases and matches GMS-DH closely in the remaining case. Remark especially how NEPF maintains feasibility for the RCTSP, which the other neural methods struggle with. Gurobi also achieves a relatively low feasibility rate, which could likely be mitigated with a warm-start approach or a more stable mixed-integer linear program formulation.  

\begin{table*}
    \centering
    \caption{Statistics for the distributions in this section.}
    \label{tab: real_stat}
    \begin{tabular}{lccc}
\toprule
& NC 100 & WC 100 & SC 100 \\
\midrule
Max $M$ per node pair & 22 & 17 &  12 \\
Mean $M$ per node pair & 4.70 & 2.80 &  1.80 \\
Mean $\#$edges per graph & 47k & 28k & 18k\\
\bottomrule
\end{tabular}
\end{table*}

\begin{table*}
    \centering
    \caption{Results across 100 instances induced by multi-objective shortest path calculation. The objective value and gap are calculated over feasible instances only.}
    \label{tab: real}
    {
    \setlength{\tabcolsep}{0.07cm}
    \renewcommand{\arraystretch}{0.8}
    \fontsize{8}{11}\selectfont
    \begin{tabular}{clcccccccccccccc}
\toprule
&\multicolumn{1}{l|}{}                        
& \multicolumn{4}{c|}{NC 100}                                              
& \multicolumn{4}{c|}{WC 100}                          
& \multicolumn{4}{c}{SC 100} \\
&\multicolumn{1}{l|}{}                  
& HV/Obj. & Gap & Feas. & \multicolumn{1}{c|}{Time}   
& HV/Obj. & Gap & Feas. & \multicolumn{1}{c|}{Time}   
& HV/Obj. & Gap & Feas. & Time   \\ 
\midrule

\multirow{9}{*}{\rotatebox{90}{\textbf{MOTSP}}} 

&\multicolumn{1}{l|}{LKH}          
& 0.74 & 0.00\% & - & \multicolumn{1}{c|}{(5.3m)}  
& 0.67 & 0.00\% & - & \multicolumn{1}{c|}{(5.7m)}
& 0.70 & 0.00\% & - & (8.2m) 
\\

&\multicolumn{1}{l|}{OR-tools}          
& 0.71 & 3.95\% & - & \multicolumn{1}{c|}{(2.8m)}  
& 0.64 & 4.16\% & - & \multicolumn{1}{c|}{(2.6m)}
& 0.69 & 2.03\% & - & (2.0m) 
\\

&\multicolumn{1}{l|}{NN}          
& 0.67 & 9.31\% & - & \multicolumn{1}{c|}{(13s)}  
& 0.60 & 10.65\% & - & \multicolumn{1}{c|}{(13s)}
& 0.65 & 8.31\% & - & (13s) 
\\

\cmidrule(lr){2-14}

&\multicolumn{1}{l|}{GMS-EB}          
& 0.45 & 39.63\% & - & \multicolumn{1}{c|}{(17m)}  
& 0.62 & 7.24\% & - & \multicolumn{1}{c|}{(13m)}
& 0.62 & 12.38\% & - & (9.1m) 
\\

&\multicolumn{1}{l|}{GMS-EB (aug)}          
& \underline{0.71} & \underline{4.29\%} & - & \multicolumn{1}{c|}{(2.2h)}  
& \underline{0.64} & \underline{4.14\%} & - & \multicolumn{1}{c|}{(1.7h)}
& \underline{0.67} & \underline{4.50\%} & - & (1.2h)
\\

&\multicolumn{1}{l|}{GMS-DH}          
& 0.64 & 13.15\% & - & \multicolumn{1}{c|}{(34s)}  
& 0.59 & 12.28\% & - & \multicolumn{1}{c|}{(29s)}
& 0.65 & 7.91\% & - &  (26s)
\\

&\multicolumn{1}{l|}{GMS-DH (aug)}          
& 0.70 & 5.63\% & - & \multicolumn{1}{c|}{(4.8m)}  
& 0.63 & 5.10\% & - & \multicolumn{1}{c|}{(4.0m)}
& 0.66 & 6.09\% & - & (3.6m)
\\

\cmidrule(lr){2-14}

&\multicolumn{1}{l|}{NEPF}          
& 0.70 & 5.17\% & - & \multicolumn{1}{c|}{(8.2s)}  
& 0.63 & 5.89\% & - & \multicolumn{1}{c|}{(7.9s)}
& 0.66 & 5.77\% & - & (7.6s) 
\\

&\multicolumn{1}{l|}{NEPF (aug)}          
& \textbf{0.71} & \textbf{3.69\%} & - & \multicolumn{1}{c|}{(50s)}  
& \textbf{0.64} & \textbf{4.04\%} & - & \multicolumn{1}{c|}{(48s)}
& \textbf{0.67} & \textbf{4.19\%} & - & (47s)
\\

\midrule \midrule

\multirow{9}{*}{\rotatebox{90}{\textbf{RCTSP}}} 

&\multicolumn{1}{l|}{Gurobi}          
& 11.9 & 0.00\% & 46\% & \multicolumn{1}{c|}{(20m)}  
& 9.97 & 0.00\% & 22\% & \multicolumn{1}{c|}{(22m)}
& 7.77 & 0.00\% & 79\% & (16m) 
\\

&\multicolumn{1}{l|}{Beam Search}          
& 17.5 & 46.89\% & 100\% & \multicolumn{1}{c|}{(13m)}  
& 15.0 & 50.37\% & 97\% & \multicolumn{1}{c|}{(10m)}
& 10.6 & 35.93\% & 100\% & (5.5m)
\\

\cmidrule(lr){2-14}

&\multicolumn{1}{l|}{GMS-EB}          
& 29.5 & 147.20\% & 94\% & \multicolumn{1}{c|}{(12s)}  
& 18.4 & 84.01\% & 48\% & \multicolumn{1}{c|}{(17s)}
& 11.4 & 46.41\% & 99\% & (6.4s)
\\

&\multicolumn{1}{l|}{GMS-EB (aug)}          
& 24.6 & 105.96\% & 100\% & \multicolumn{1}{c|}{(1.6m)}  
& 18.2 & 82.90\% & 69\% & \multicolumn{1}{c|}{(2.2m)}
& 9.92 & 27.71\% & 100\% & (48s) 
\\

&\multicolumn{1}{l|}{GMS-DH}          
& 18.5 & 54.78\% & 100\% & \multicolumn{1}{c|}{(3.5s)}  
& \underline{14.6} & \underline{45.97\%} & 37\% & \multicolumn{1}{c|}{(2.5s)}
& 9.94 & 27.90\% & 100\% & (1.8s) 
\\

&\multicolumn{1}{l|}{GMS-DH (aug)}          
& \textbf{17.1} & \textbf{43.36\%} & 100\% & \multicolumn{1}{c|}{(30s)}  
& 15.0 & 50.31\% & 89\% & \multicolumn{1}{c|}{(18s)}
& \underline{9.60} & \underline{23.57\%} & 100\% & (13s) 
\\

\cmidrule(lr){2-14}

&\multicolumn{1}{l|}{NEPF}          
& 18.3 & 53.83\% & 100\% & \multicolumn{1}{c|}{(1.0s)}  
& 14.0 & 40.11\% & 98\% & \multicolumn{1}{c|}{(0.9s)}
& 9.54 & 22.73\% & 100\% & (0.7s) 
\\

&\multicolumn{1}{l|}{NEPF (aug)}          
& \underline{17.4} & \underline{46.03\%} & 100\% & \multicolumn{1}{c|}{(5.4s)}  
& \textbf{13.4} & \textbf{34.79\%} & 100\% & \multicolumn{1}{c|}{(4.4s)}
& \textbf{9.24} & \textbf{18.86\%} & 100\% & (3.9s) 
\\

\bottomrule
\end{tabular}
    }
\end{table*}

\subsection{Real-World Instances}
\label{app: real-world-instances}
We complement the analysis in the previous section with trained results on instances derived from real road networks. In particular, we consider the MOTSP with varying instance sizes ($N = 20, 50, 100$). We let one objective be the travel time and the other energy consumption. Both quantities are derived with the widely used model of \citet{BASSO2019141}, which takes into account speed profiles, elevation changes, intersection stops and vehicle data. By using this realistic model we ensure there is a real trade-off between the edge features, since the fastest path tends to not be the most energy efficient one, motivating the multigraph framework. We then sample nodes uniformly in the San Francisco road network (obtained via OpenStreetMap) and calculate multigraph edges via a multi-objective shortest path procedure, capping alternative paths to 5 at most. We train one model for each instance size, with the same setup as the main experiments. 

These results are displayed in Table~\ref{tab: sf_results}. NEPF (aug) achieves the best neural performance on two of the three sizes ($N=20$ and $N=100$) and is within $0.4\%$ HV of GMS-DH (aug) on the third, while being substantially faster than all baselines at every size. The runtime advantage grows with instance size. At $N=100$, NEPF requires $11$s compared to $39$s for GMS-DH and $7.6$m for GMS-EB ($1.4$m vs.\ $5.4$m vs.\ $1.0$h with augmentation). Notably, GMS-EB degrades sharply at $N=100$ on these instances ($0.7561$ HV with augmentation, a $20.8\%$ gap). LKH remains the overall quality reference, but at roughly an order of magnitude higher runtime ($14.6$m vs.\ $1.4$m at $N=100$).

Overall, these results confirm that the strong performance of NEPF transfers to instances with realistic structure, including variable numbers of parallel edges, correlation induced by a physical trade-off between travel time and energy consumption and metric properties inherited from the real road network. 

\begin{table}[t]
\centering
\caption{Results on real-world instances derived from the San Francisco road network,
with energy consumption and travel time as objectives (MOTSP, $M \leq 5$ parallel
edges per node pair, $N \in \{20, 50, 100\}$).}
\label{tab: sf_results}
\setlength{\tabcolsep}{3.5pt}
\small
\begin{tabular}{lccccccccc}
\toprule
& \multicolumn{3}{c}{SF5-20} & \multicolumn{3}{c}{SF5-50} & \multicolumn{3}{c}{SF5-100} \\
\cmidrule(lr){2-4}\cmidrule(lr){5-7}\cmidrule(lr){8-10}
 & HV & Gap & Time & HV & Gap & Time & HV & Gap & Time \\
\midrule
LKH          & 0.63 & 0.00\% & (22s) & 0.82 & 0.00\% & (2.6m) & 0.95 & 0.00\% & (15m) \\
OR-Tools     & 0.59 & 6.25\% & (6.3s)  & 0.73 & 11.46\% & (43s) & 0.77 & 19.08\% & (4.0m) \\
\midrule
GMS-EB       & 0.62 & 1.77\% & (4.0s)  & 0.78 & 4.81\% & (1.0m) & 0.73 & 23.78\% & (7.6m) \\
GMS-EB (aug) & 0.62 & 1.40\% & (30s) & 0.78 & 4.27\% & (8.2m) & 0.76 & 20.79\% & (1.0h) \\
GMS-DH       & 0.62 & 1.94\% & (2.2s)  & 0.79 & 3.45\% & (9.0s) & 0.90 & 5.89\% & (39s) \\
GMS-DH (aug) & \underline{0.62} & \underline{1.39\%} & (10s) & \textbf{0.80} & \textbf{2.75\%} & (1.1m) & \underline{0.91} & \underline{5.19\%} & (5.4m) \\
NEPF         & 0.62 & 1.83\% & (1.2s)  & 0.79 & 3.66\% & (3.1s) & 0.90 & 5.68\% & (11s) \\
NEPF (aug)   & \textbf{0.62} & \textbf{0.97\%} & (2.0s) & \underline{0.79} & \underline{3.11\%} & (15s) & \textbf{0.91} & \textbf{5.04\%} & (1.4m) \\
\bottomrule
\end{tabular}
\end{table}

\subsection{Further Ablations}
\label{app: further_ablations}
This section ablates further components of our framework, beyond those investigated in Section~\ref{sec: ablations}. Throughout, we utilize the MOTSPTW on FLEX2-100 for comparisons.

\paragraph{Half-precision inference.} Inference in full precision increases the total runtime (for solving 200 instances) to circa $32$s from $17$s, while not impacting solution quality. Remark that all neural baselines utilize half-precision inference in our experiments, ensuring a fair runtime comparison. 

\paragraph{FSASP samples.} Figure~\ref{fig: K2} shows the effect of changing the number of FSASP samples. While increasing the number of samples only moderately impacts the solution time, our inference choice $K_2 = 50$ balances runtime vs. hypervolume. Also remark that already at $K_2 = 1$ NEPF produces high-quality solutions, which indicates that the FSASP module learns a sensible distribution over the multigraph edges. 

\begin{figure}
    \centering
    \definecolor{classicblue}{RGB}{72,120,168}
\definecolor{nepforange}{RGB}{232,133,61}
\definecolor{nepfred}{RGB}{192,57,43}

\begin{tikzpicture}[font=\footnotesize]
\begin{axis}[
  width=7.2cm, height=4.8cm,
  xmode=log,
  log ticks with fixed point,
  xtick={1,2,5,10,20,50,100},
  xmin=0.8, xmax=130,
  xlabel={FSASP samples $K_2$},
  ylabel={HV ($\uparrow$)},
  ymin=0.908, ymax=0.933,
  grid=both,
  grid style={line width=0.2pt, draw=gray!25},
  tick label style={font=\footnotesize},
  label style={font=\footnotesize},
  ytick pos=left,
  every axis plot/.append style={line width=0.9pt},
  extra description/.code={%
    \draw[black, line width=0.4pt]
      (axis description cs:1,0) -- (axis description cs:1,1);
    \foreach \t/\f in {14/0, 16/0.25, 18/0.5, 20/0.75, 22/1}{%
      \draw[black, line width=0.4pt]
        (axis description cs:1,\f) -- ++(5pt,0)
        node[right, xshift=2pt, inner sep=1.5pt, font=\footnotesize] {\t};}
    \node[rotate=90, anchor=south, font=\footnotesize]
      at ([xshift=32pt]axis description cs:1,0.5) {Time (s)};
  },
]
\addplot[classicblue, mark=*, mark size=1.6pt] coordinates
  {(1,0.9124) (10,0.9231) (20,0.9261) (50,0.9282) (100,0.9294)};

\addplot[nepfred, dashed, mark=square*, mark size=1.4pt,
         mark options={solid, fill=nepfred}] coordinates
  {(1,0.9135) (10,0.914250) (20,0.9160) (50,0.917375) (100,0.926750)};

\node[classicblue, font=\tiny, anchor=south east, inner sep=1.5pt]
  at (axis cs:4,0.9190) {HV};
\node[nepfred, font=\tiny, anchor=north west, inner sep=1.5pt]
  at (axis cs:4,0.9138) {Time};

\draw[dashed, gray, line width=0.6pt]
  (axis cs:20,0.908) -- (axis cs:20,0.933)
  node[pos=0.94, left, font=\tiny, gray] {Training};
\draw[dashed, nepforange, line width=0.6pt]
  (axis cs:50,0.908) -- (axis cs:50,0.933)
  node[pos=0.06, right, font=\tiny, nepforange] {Inference};
\end{axis}
\end{tikzpicture}
    \caption{Impact of FSASP samples on solution quality (HV) and total runtime.}
    \label{fig: K2}
\end{figure}

\paragraph{Curriculum learning.} Training without curriculum learning (described in Appendix~\ref{app: hyper}) yields hypervolume 0.8599 on the test set, which should be compared to 0.9282 with curriculum. At the same time, the hypervolume on the validation set plateaus after circa 10 epochs, indicating that the model reaches a local minimum. The curriculum seems especially important in the multi-objective setting, where solution diversity across preferences tends to collapse when the models are trained directly on large instance sizes.  

\paragraph{Max reward.} Section~\ref{sec: hierarchical_training} establishes $R^*(\boldsymbol{\pi}_{ij}) = \max_{k=1, \dots, K_2} R(\boldsymbol{\pi}_{ij}, \boldsymbol{\epsilon}_{ijk})$ as the reward for the node permutation stage. We argue that this is a sensible choice, as the quality of a node permutation should be evaluated by the best possible downstream edge selection. However, an alternative is to consider the mean reward:
\begin{equation}
    R^*(\boldsymbol{\pi}_{ij}) = \frac{1}{K_2} \sum_{k=1}^{K_2} R(\boldsymbol{\pi}_{ij}, \boldsymbol{\epsilon}_{ijk}).
\end{equation}
Using this option for training, we obtain the test hypervolume 0.9172 for the MOTSPTW on FLEX2-100, a noticeable degradation from the baseline 0.9282 with the max-reward.

\end{document}